\documentclass[sigplan,nonacm]{acmart}

\usepackage{xspace}
\usepackage{enumitem}
\usepackage{siunitx}
\usepackage{colortbl}
\usepackage{listings}
\usepackage{xcolor}

\NewDocumentCommand{\var}{O{s} m O{}}{%
  \ensuremath{#1_{#2}^{#3}}% add \vphantom{<bizarre sup>}
}

\usepackage{pifont}
\newcommand{\cmark}{\ding{51}}%
\newcommand{\xmark}{\ding{55}}%

\newcommand{\commentout}[1]{}

\definecolor{light-gray}{gray}{0.80}

\newcommand\fref{Fig.~\ref}
\newcommand\tref{Table~\ref}
\newcommand\sref{\S~\ref}
\usepackage{amsthm}

\usepackage{amsthm}

\newtheorem{mytheorem}{Theorem}
\newtheorem{assumption}[mytheorem]{Assumption}
\newtheorem{lemma}[mytheorem]{Lemma}
\newtheorem{remk}[mytheorem]{Remark}

\newcommand{\superoffload}{SuperOffload\xspace}
\newcommand{\superchip}{Superchip\xspace}
\newcommand{\superchips}{Superchips\xspace}
\newcommand{\zerooffload}{ZeRO-Offload\xspace}

\newcommand{\scdfg}{SA-DFG\xspace}
\newcommand{\superulysses}{SuperOffload-Ulysses\xspace}

\begin{document}
\title{SuperOffload: Unleashing the Power of Large-Scale LLM Training on Superchips} 

\author{Xinyu Lian}
\affiliation{%
  \institution{SSAIL Lab, University of Illinois}
  \city{Urbana-Champaign}
  \state{IL}
  \country{USA}
}

\author{Masahiro Tanaka}
\affiliation{%
  \institution{Anyscale}
  \city{San Francisco}
  \state{CA}
  \country{USA}
}

\author{Olatunji Ruwase}
\affiliation{%
  \institution{Snowflake}
  \city{Bellevue}
  \state{WA}
  \country{USA}
}

\author{Minjia Zhang}
\affiliation{%
  \institution{SSAIL Lab, University of Illinois}
  \city{Urbana-Champaign}
  \state{IL}
  \country{USA}
}
  
\begin{abstract}

The emergence of \superchips represents a significant advancement in next-generation AI hardware. These \superchips employ a tightly coupled heterogeneous architecture that integrates GPU and CPU on the same package, which offers unprecedented computational power. However, there has been scant research investigating how LLM training benefits from this new architecture.
In this work, for the first time, we study LLM training solutions based on offloading for \superchips.
% using NVIDIA's Grace Hopper Superchip (GH200). 
We observe important differences between \superchips and traditional loosely-coupled GPU-CPU architecture, which necessitate revisiting prevailing assumptions about offloading. Based on that, we present \superoffload, a \superchip-centric offloading system that simultaneously uses Hopper GPU, Grace CPU, and NVLink-C2C interconnect more efficiently. \superoffload accomplishes this via a combination of techniques, such as adaptive weight offloading, bucketization repartitioning, \superchip-aware casting, speculative execution, and a highly optimized Adam optimizer for Grace CPUs. Our evaluation of \superoffload on NVIDIA GH200 demonstrates up to 2.5$\times$ throughput improvement compared to state-of-the-art offloading-based systems, enabling training of up to 25B model on a single \superchip while achieving high training throughput.
We also extend \superoffload with ZeRO-style data parallelism and DeepSpeed-Ulysses sequence parallelism, enabling training of 13B model with sequence lengths up to 1 million tokens on 8 GH200 while achieving 55\% MFU. 

\end{abstract}

\maketitle % should come after the abstract

\section{Introduction}
\label{sec:intro}

The impressive capabilities of Large Language Models (LLMs) have spurred significant research and industry
efforts~\cite{gpt4,googlegemini,llama3TechReport,phi3,grok} to optimize the resource and time costs of training state-of-the-art LLMs. A large fraction of this work is motivated by the \emph{memory wall} challenge of scaling LLMs~\cite{memory-wall} -- the disparity between exponential growth in model sizes and relatively slower increase in hardware memory capacity and bandwidth~\cite{scaleing-law-nlp,zero-infinity}. Consequently, training LLMs is often constrained by limited GPU memory. 

When LLM training exceeds a single GPU's memory capacity, practitioners often employ a variety of distributed techniques, such as ZeRO-style data parallelism~\cite{zero}, tensor-slicing parallelism~\cite{megatron-lm}, pipeline parallelism~\cite{gpipe,1f1b,megatron-lm-v2}, and sequence parallelism~\cite{ulysses,ring-attention}, to shard model and data states across multiple GPUs. However, the large number of GPUs required by these methods poses a significant bottleneck for researchers and institutions with modest GPU budgets.

Yet, enabling training of LLMs with a limited number of GPUs is critical in many scenarios. Beyond pre-training, LLMs often reach their full potential through additional post-training phases, such as long context extension~\cite{llama3TechReport,chen2024longlora}, continuous training~\cite{medpalm,codellama,lian2025large}, supervised fine-tuning~\cite{sft}, instruction tuning~\cite{instruct-tuning}, and human preference alignment techniques such as RLHF~\cite{rlhf} and DPO~\cite{dpo}, all of which require model training, even when users have significantly fewer GPUs than used for pre-training. 

Among different approaches, offloading techniques have become a promising solution to reduce GPU requirements for large-scale LLM training~\cite{zero-offload,zero-infinity,dos,speedloader}. These methods offload a portion of the model states and computation from GPU to CPU, effectively utilizing both to enable large-scale LLM training without excessive GPU usage. As a result, offloading is supported by major DL training frameworks such as PyTorch-FSDP~\cite{fsdp} and DeepSpeed~\cite{zero-offload,zero-infinity}.

While showing promising results, existing offloading solutions are based on one key assumption: the GPU is an accelerator connected to the CPU via PCIe, which has limited communication bandwidth (e.g., 32GB/s for PCIe-Gen4).
Due to this bandwidth limitation, prior work primarily focuses on optimizing data transfers between GPU and CPU to avoid PCIe becoming a major performance bottleneck~\cite{fsdp-offload,zero-offload,zero-infinity}. However, hardware vendors have embarked on the development of a new class of high-end tightly coupled hardware architecture in their product roadmaps~\cite{gb200}, e.g., NVIDIA Grace Hopper GH200~\cite{grace-hopper} and AMD MI300A~\cite{amd-mi300a}, which herald a paradigm shift in deep learning infrastructure. 

Take NVIDIA's Grace Hopper Superchip (GH200) as an example. It integrates a Hopper H100 GPU and an ARM-based Grace CPU on the same package through the NVLink-Chip2Chip (C2C) interconnect, with GPU-CPU bandwidth reaching a staggering 900 GB/s—30$\times$ increase over traditional PCIe connections. Moreover, multiple Superchips can be connected to form a large-scale tightly coupled system, which offers fast and low latency interaction between different classes of chiplets. 
Given the tightly coupled hardware architecture, recent work has started benchmarking Superchips~\cite{superchip-data-study,superchip-benchmark,superchip-study} and how to best program them~\cite{pie}. However, efficient utilization of Superchips for offloading-based LLM training has not yet been studied in depth.

In this work, we investigate software optimizations to train LLMs in the era of Superchips. We find that existing offloading methods, designed under the assumption of limited PCIe bandwidth, perform suboptimally on Superchips. In particular, off-the-shelf solutions severely underutilize the Hopper GPU, Grace CPU, and C2C bandwidth. For example, prior work often adopts a greedy algorithm for data transfer in mixed-precision training, transferring only low-precision data between CPU and GPU to minimize PCIe communication volume in order to maximize training efficiency. Our detailed analysis reveals that these common assumptions do not hold true for Superchips. 

\vspace{5pt}
\noindent
\textbf{Contribution.} This paper addresses a pressing need to harness the power of Superchips for training LLMs. Concretely, we take a leap forward and introduce \superoffload, a Superchip-centric offloading system that delivers excellent training performance on NVIDIA GH200 Superchips. \superoffload accomplishes this via a combination of techniques, such as \emph{adaptive weight-stationary and weight-flow offloading} (\sref{subsec:weight-placement}), \emph{fine-grained bucketization repartitioning} (\sref{subsec:bucketization}), \emph{speculation-then-validation} support to overlap compute and communication (\sref{subsec:speculation-schedule}), 
\emph{Superchip-aware typecasting for mixed-precision training} (\sref{subsec:casting-cost}), 
and \emph{highly optimized GraceAdam} for Grace Arm CPUs  (\sref{subsec:graceadam}). To our knowledge, this is the first work that systematically optimizes LLM training for \superchip architectures, maximizing the utilization of Hopper GPU, Grace CPU, and C2C interconnect, simultaneously.

We validate our design through comprehensive experiments. \superoffload achieves a 2.5$\times$ throughput improvement over state-of-the-art offloading solutions and outperforms GPU-only approaches across all tested model sizes, which challenges the conventional wisdom that offloading often comes at the cost of performance penalty.
\superoffload also enables 25B-parameter model training on a single Superchip, which is 7$\times$ larger than GPU-only solutions.

In addition, we validate our design by integrating \superoffload with ZeRO-style data parallelism~\cite{zero}, which allows \superoffload to extend to multi-Superchip training settings. Specifically, \superoffload enables LLM training with 50B parameters using only four Superchips, which is 2.5$\times$ larger than the largest model trainable with ZeRO-Offload and FSDP-Offload, while achieving up to 3$\times$ higher training throughput.

\begin{figure}[!t]
    \centering
    \includegraphics[width=0.95\linewidth]{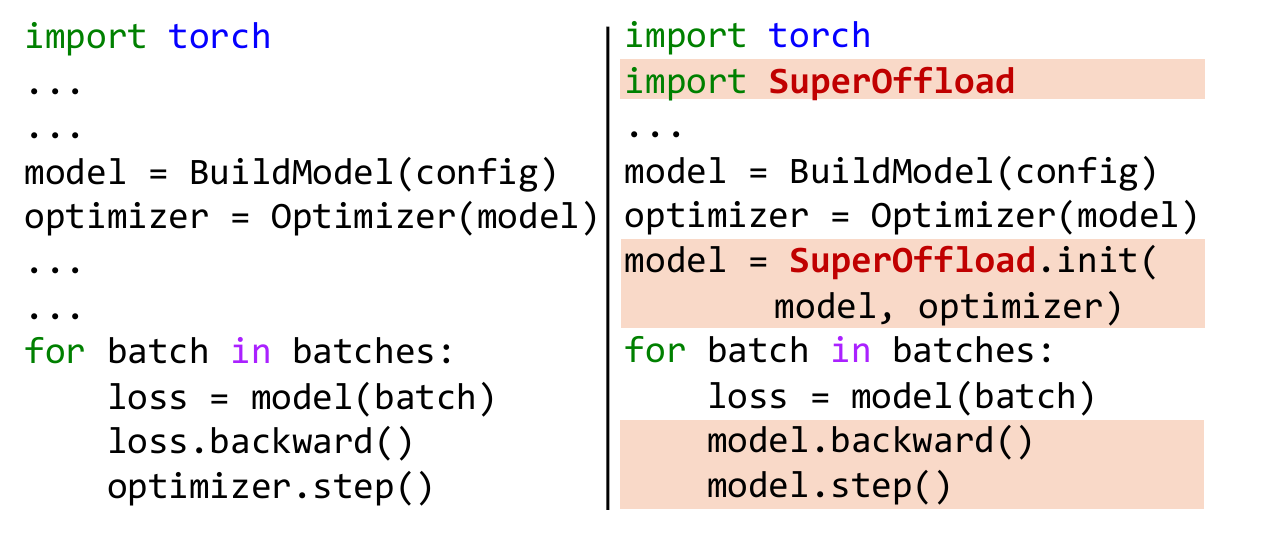}
    \caption{\superoffload can be enabled with a few lines of change. The code on left shows a standard training pipeline, while the right shows the same pipeline with \superoffload.}
    \label{fig:code}
\end{figure}

We also extend \superoffload to support long-sequence training, termed \superulysses, via integration with Ulysses-style sequence parallelism~\cite{ulysses}. This unlocks new possibilities for training or fine-tuning LLMs with long sequence lengths. Specifically, \superulysses supports training sequences 8$\times$ longer than Ulysses and {enables the training of a 13B model with up to a million tokens using only 8 GH200 Superchips, while achieving 55\% MFU}. 
To enhance usability, \superoffload has been integrated with DeepSpeed~\cite{deepspeed}, a popular open-source DL training library. Users can enable \superoffload through a few lines, as illustrated in \fref{fig:code}, {without model code refactoring}.

\section{Background and Related Work}
\label{sec:background}

\subsection{The Emergence of Superchips}

\begin{figure}[t]
    \centering
    \includegraphics[width=\linewidth]{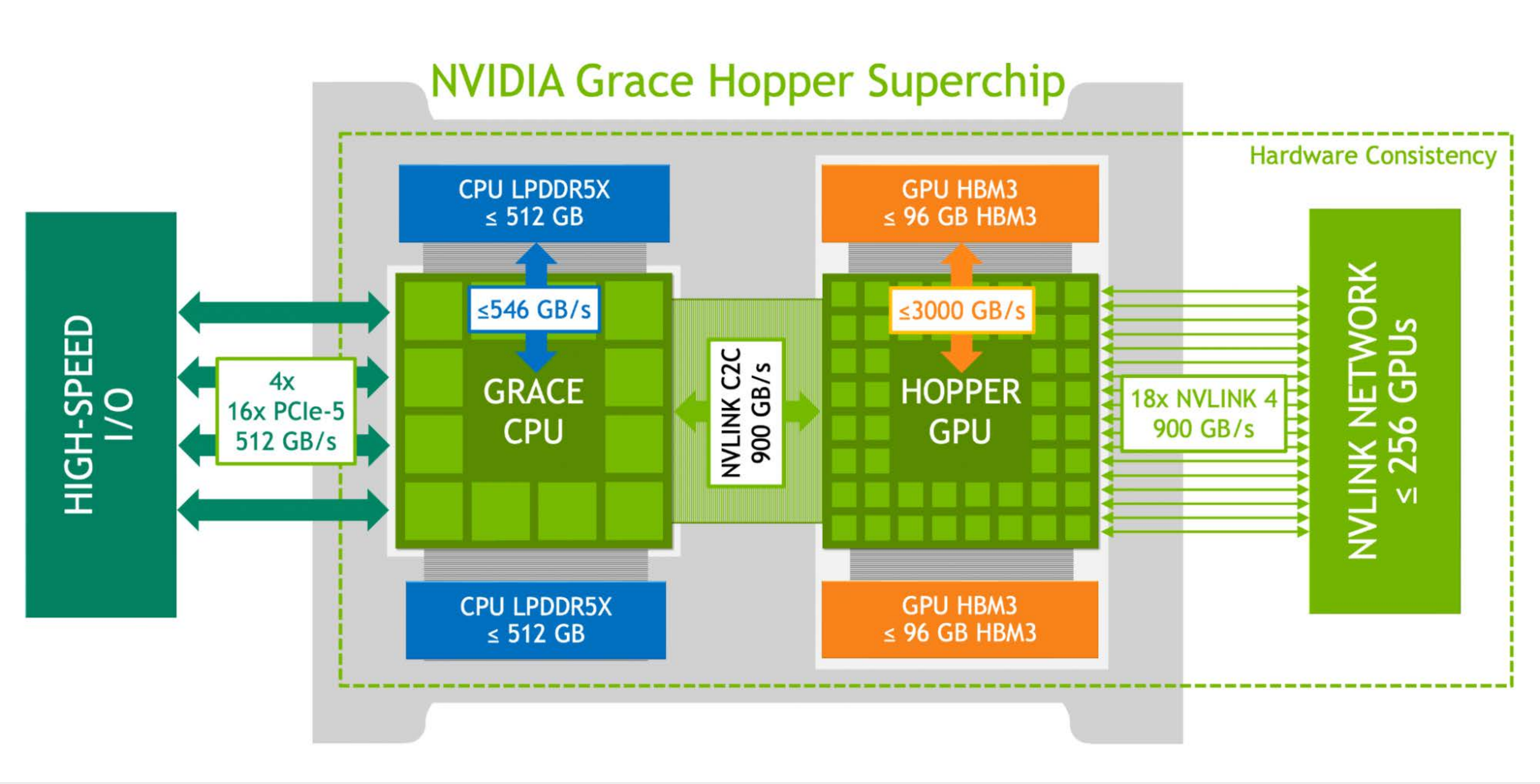}
    \caption{Grace Hopper GH200 Superchip hardware overview~\cite{grace-hopper-architecture}. The Grace CPU, equipped with a 500\,GB/s DDR interface, is shown on the left, and the Hopper GPU, with 4000\,GB/s HBM, is depicted on the right. The two are interconnected via NVLINK-C2C at 900\,GB/s.}
    \label{fig:GH-overview}
\end{figure}

GPUs were initially developed as accelerators loosely-coupled with CPUs. 
Driven by the need to support the next generation of AI systems, NVIDIA introduced GH200 Grace Hopper Superchip~\cite{grace-hopper}, a groundbreaking processor that integrates a tightly-coupled Hopper GPU and Grace CPU on the same package with high-bandwidth NVLink Chip-2-Chip (C2C) interconnect, as shown in \fref{fig:GH-overview}. Recently, NVIDIA also announced GB200~\cite{gb200}, the next-generation Superchip that has a tightly-coupled Blackwell GPU with Grace CPU to deliver unprecedented performance and efficiency for AI and other advanced computing fields. Other hardware vendors, such as AMD, have released AMD Instinct MI300A Superchip~\cite{amd-mi300a} that connects 6 AMD CDNA3 GPU chiplets and 3 AMD Zen4 x86 CPU chiplets through AMD's 4th Gen Infinity interconnect using 3D stacking technology. 

Despite being a new architecture, Superchips have already been adopted in large-scale supercomputing clusters worldwide, including France's EXA1-HE~\cite{exa1-he}, Poland's Helios~\cite{helios}, Switzerland's Alps~\cite{alps}, Germany's JUPITER~\cite{jupiter}, the United States' NCSA~\cite{ncsa-deltai,cui2025delta}, and Japan's Miyabi~\cite{miyabi}. Notably, the Alps supercomputer has integrated over 2,000 GH200 nodes and Isambard-AI supercomputer plans to deploy over 5,000 GH200 nodes to enhance computational capabilities. The GH200 Superchip is also available through cloud service providers such as Amazon Web Services~\cite{aws-gh200}, Microsoft Azure~\cite{azure-gh200}, and Lambda~\cite{lambda-gh200}, offering flexible  options for organizations seeking high-performance computing resources.

Given the ever-growing adoption, Superchips have attracted significant attention with active research on benchmarking their hardware characteristics~\cite{superchip-data-study,superchip-benchmark,superchip-study} and their programmability~\cite{pie}. However, few studies have been done to investigate how to effectively leverage it for LLM training.

\subsection{LLM Training on Heterogeneous Hardware}

Modern LLM training is highly memory-intensive due to the large number of parameters and the significant memory required for model states like gradients and optimizer states~\cite{adam, adamw}. As an example, during mixed precision training~\cite{mixed-precision-training}, a model with $\Psi$ parameters consumes
a total of 16$\Psi$ bytes of memory (e.g., 2$\Psi$ parameters, 2$\Psi$ gradients, and 12$\Psi$ optimizer states). An NVIDIA H100 GPU with 96GB memory can only accommodate a model with up to 6B parameters.

To overcome the memory limitations, advanced distributed training techniques such as tensor parallelism (TP)~\cite{megatron-lm}, pipeline parallelism (PP)~\cite{gpipe,1f1b}, ZeRO-style data parallelism (ZeRO-DP)~\cite{zero}, 3D parallelism~\cite{megatron-lm-v2} have been developed.  
Despite their strong capabilities, all of these methods rely on the aggregated memory of multiple GPUs to store the model states and residual states. 

To reduce the number of GPUs needed, researchers have explored heterogeneous training that exploit the expansive memory capacity of CPUs to scale up model training~\cite{autotm,swapadvisor,capuchin,sentinel,vdnn,superneurons}. 
% Several prior research efforts have explored offloading strategies to improve memory efficiency for LLM training~\cite{zero-offload,zero-infinity,dos,speedloader}.
Offloading strategies like ZeRO-Offload~\cite{zero-offload} carefully divide the computation between the hardware types: GPU performs compute-intensive operations such as backpropagation, while CPU performs memory-intensive operations such as optimizer updates. ZeRO-Infinity further extends this approach to NVMe storage, enabling larger model training with limited GPU resources~\cite{zero-infinity}.
Deep-Optimizer-States extends \zerooffload by fetching optimizer states from CPU to GPU and updating parameters in parallel across both devices, thus reducing optimizer step time in the critical path~\cite{dos}. Similar to ZeRO-Infinity, SpeedLoader offloads both parameters and optimizer states to CPU memory, then dynamically fetches them to GPU layer by layer~\cite{speedloader}. 

While demonstrating promising results, prior work often assume limited bandwidth between GPU and CPU via PCIe (e.g., $<$32GB/s). 
Therefore, they primarily focus on reducing the communication volume between GPU and CPU to avoid the PCIe from becoming a major performance bottleneck. 
However, newer hardware, such as Superchips with 30$\times$ higher C2C bandwidth, challenges these assumptions. Consequently, adhering to the older design principles could lead to suboptimal performance, which motivates us to revisit the offloading design.

\section{Challenges and Opportunities}
\label{sec:opportunity}

NVIDIA's GH200 Grace Hopper Superchip represents a significant advancement in high-performance computing. However, this new hardware architecture poses challenges for existing off-the-shelf offloading solutions to \emph{fully utilize all Superchip resources, including the Hopper GPU, C2C bandwidth, simultaneously}, for the following reasons:

\begin{table}[!t]
\setlength{\tabcolsep}{2.5pt}
\renewcommand{\arraystretch}{1.1}
\centering
\caption{Comparison of GPU node. DGX-2 is used in the ZeRO-Offload~\cite{zero-offload}, DGX-A100 is used for LLaMa model training~\cite{llama}, and GH denotes a single Grace Hopper Superchip.}
{\footnotesize
\begin{tabular}{lccc}
\toprule
\textbf{Node Arch} & \textbf{DGX-2} & \textbf{DGX-A100} & \textbf{GH} \\ 
Hardware Setting & Intel Xeon+V100 & AMD Roma+A100 & GH200 \\
\midrule
% DGX-V: DDR4-2666, 6 memory channels
% DGX-A: DDR4‑3200, 8 memory channels
CPU BW (GB/s) & 100 & 150 & \textbf{500} \\
C$\leftrightarrow$GPU BW (GB/s) & 32 & 64 & \textbf{900} \\
CPU Cores & 24 & 64 & 72 \\ 
CPU FLOPS (TFLOPS)& 2.07 & 2.3  & 3.0 \\ 
GPU FLOPS (TFLOPS)& 125.0 & 312.0 & 990.0 \\ 
GPU/CPU FLOPS & 60.39 & 135.65 & \textbf{330.0} \\
% C$\leftrightarrow$G BW to FLOPS ratio & 7.14 & 3.28 & 17.65 \\ \hline
\bottomrule
\end{tabular}}
\label{tab:node_arch_comparison}
\end{table}

\paragraph{Superchip $\neq$ GPU + CPU} 
Before the advent of Superchip, the connection between CPUs and GPUs relied on PCI Express (PCIe). 
For instance, as shown in \tref{tab:node_arch_comparison}, the DGX-A100 node~\cite{dgx-a100}, widely deployed within Meta for training the LLaMA model~\cite{llama}, comprises two AMD Rome CPUs and eight NVIDIA A100 GPUs, connected via PCIe 4.0 x16, delivering a bandwidth of 64 GB/s. Similarly, the DGX-2 node evaluated in ZeRO-Offload~\cite{zero-offload} contains two Intel Xeon CPUs and 16 NVIDIA V100 GPUs, with the CPU-GPU connection utilizing PCIe 3.0 x16, providing a bandwidth of 32 GB/s.

The Superchip, however, tightly couples GPU and CPU with in the same package through the high-bandwidth interconnect, delivering a staggering 900 GB/s bandwidth between GPU and CPU (NVLINK-C2C as in the GH200) -- 14$\times$ the standard PCIe Gen4 lanes. 
In addition, the 72-core Grace CPU significantly enhances compute capacity with 500 GB/s of LPDDR5 high-bandwidth memory, providing not only additional memory space but also efficient handling of computation tasks.

\begin{figure}[!t]
    \centering
    \includegraphics[width=\linewidth]{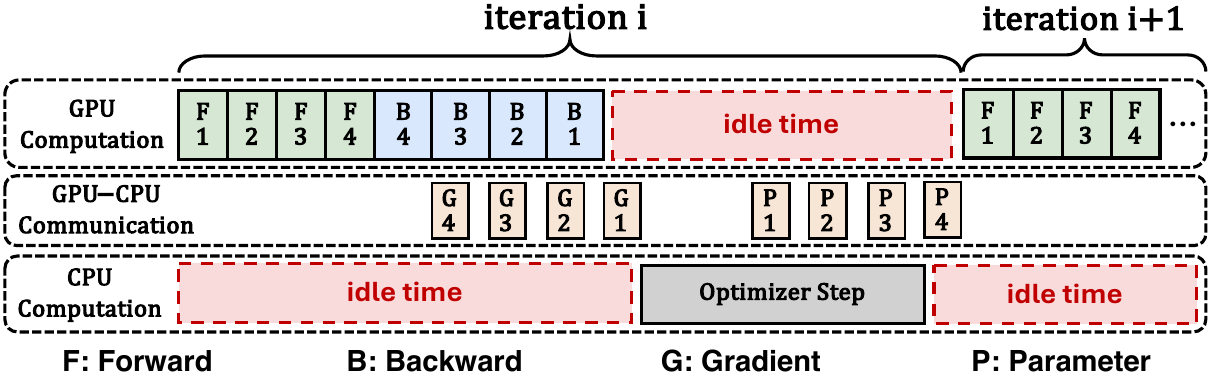}
    \caption{ZeRO-Offload computation idle time on both GPU and CPU side.}
    \vspace{-10pt}
    \label{fig:zero-offload_schedule}
\end{figure}

\begin{figure}[t]
    \centering
    \includegraphics[width=\linewidth]{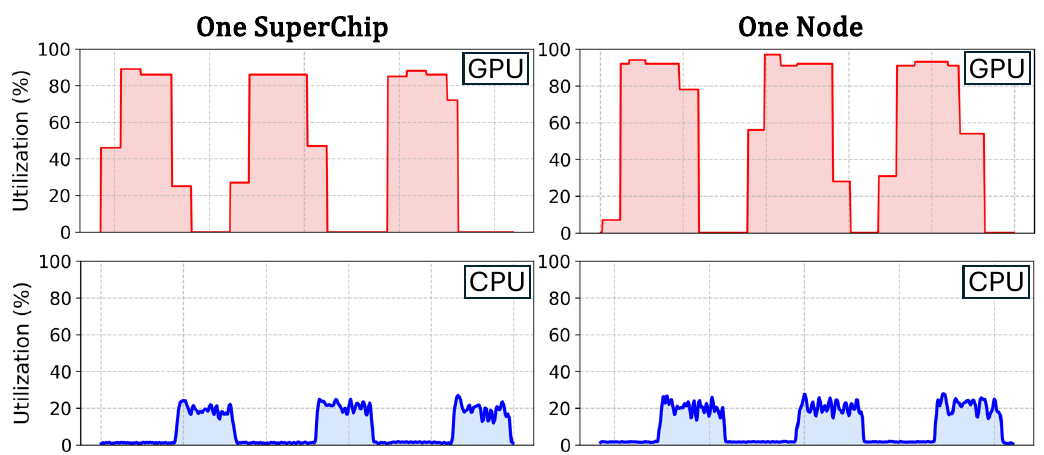}
    \caption{Prior offloading-based solutions cause idle time on both GPU and CPU side. We evaluate it using the largest model size that offloading-based solutions (ZeRO-Offload) can accommodate, along with the maximum batch size that prevents out-of-memory errors, on single \superchip and one node. The results show that during each training iteration, the GPU remains idle for 40-50\% of the total execution time.}
    \label{fig:idle}
\end{figure}

\paragraph{Challenges in idle-free scheduling.}
Prior offloading-based solutions often suffer from GPU and CPU idle time due to GPU-CPU synchronization.  
Take ZeRO-Offload~\cite{zero-offload} as an example. As shown in \fref{fig:zero-offload_schedule}, there are three primary constraints:
1) Global Gradient Synchronization:
Gradient clipping and normalization require global information. Consequently, the CPU must wait until it has received all gradients from the GPUs before starting the optimizer step. This results in CPU idle time while the GPUs are busy with the backward pass.
2) Synchronized Parameter Updates:
The GPU must wait for all FP16 parameters to return before initiating the forward pass of the next iteration. This synchronization is crucial to ensure that the forward pass uses updated parameters.
3) Limited Overlap of Computation and Communication:
Even with small bucket sizes for transferring gradients to the CPU and updated parameters back to the GPU, complete overlap between computation and communication is unattainable (more details in \sref{subsec:bucketization}). 
 As a result, 
 during the final bucket's transfer interval, 
 both the GPU and CPU remain idle.
\fref{fig:idle} shows that 40-50\% of the Hopper GPU is idle per training iteration. 
These significant idle periods raise the question: 
How can we reduce these inefficiencies for offloading-based solutions on Superchips?

\paragraph{Challenges in mixed-precision training on Superchips.}
Modern DL training often employs mixed precision training~\cite{mixed-precision-training}, which uses a mix of full and lower precisions to deliver significant computational speedup by executing operations in a lower precision format on matrix cores (e.g., TensorCore) as much as possible, while storing critical information in the full-precision to preserve model accuracy.

Mixed precision training makes offloading more complex because of additional casting operations in the computation graph. 
To enable efficient mixed-precision training through offloading, the conventional design principle was to adopt a greedy edge-cut algorithm~\cite{zero-offload}.
As an example, ZeRO-Offload keeps all the optimizer states on the CPU, and moves the FP16 gradients to CPU during the backward pass and returns the updated FP16 parameter back to GPU during the optimizer step. This
minimizes the communication volume between the GPU and CPU (e.g., 4$\Psi$), 
such that the PCIe bandwidth does not become a major bottleneck. 
However, our detailed performance analysis shows that this method is suboptimal for Superchip architectures (\sref{subsec:casting-cost}), which motivates us to revisit the decision of offloading-based mixed precision training. 

There are other challenges such as Grace Arm CPU's lack of a high-performance Adam optimizer, and lacking of a clear solution towards scaling model sizes and long-sequence on multiple Superchips. In the following, we describe the design of \superoffload, which specifically addresses all of the above challenges, eventually leading to a highly-optimized solution tailored for Superchips in many practical settings.  

\section{\superoffload Design and Implementation}
\label{sec:design}

\subsection{System Overview}
\label{subsec:overview}

% Prior work uses data-flow graphs (DFGs) to model DL training computations, where vertices represent tensor operators and edges represent data dependencies and formulate offloading as a bi-way partitioning problem~\cite{zero-offload}. A variety of heuristics have been proposed by \cite{zero-offload} to address the problem. In particular, it places optimizer updates on CPU and adopts a greedy edge-cut algorithm to minimize the communication between GPU and CPU. While being an effective approach for the conventional loosely-coupled GPU-CPU architecture, this method does not consider the Superchip architecture, leading to suboptimal performance. 

\superoffload takes the tightly-coupled \superchip architecture into account by considering a Superchip-aware DFG (\scdfg). In \scdfg, each vertex $u$ still represents a tensor operator but it also reflects the computation cost that $u$ would incur if it is assigned to the Hopper GPU or Grace CPU.
The edge $u \rightarrow v$ in \scdfg reflects the communication cost that $u \rightarrow v$  would incur if $u$ and $v$ are executed on a Hopper GPU or Grace CPU. An offload strategy then can be viewed as a two-way partition of \scdfg.  

\begin{figure}[t]
    \centering
    \includegraphics[width=\linewidth]{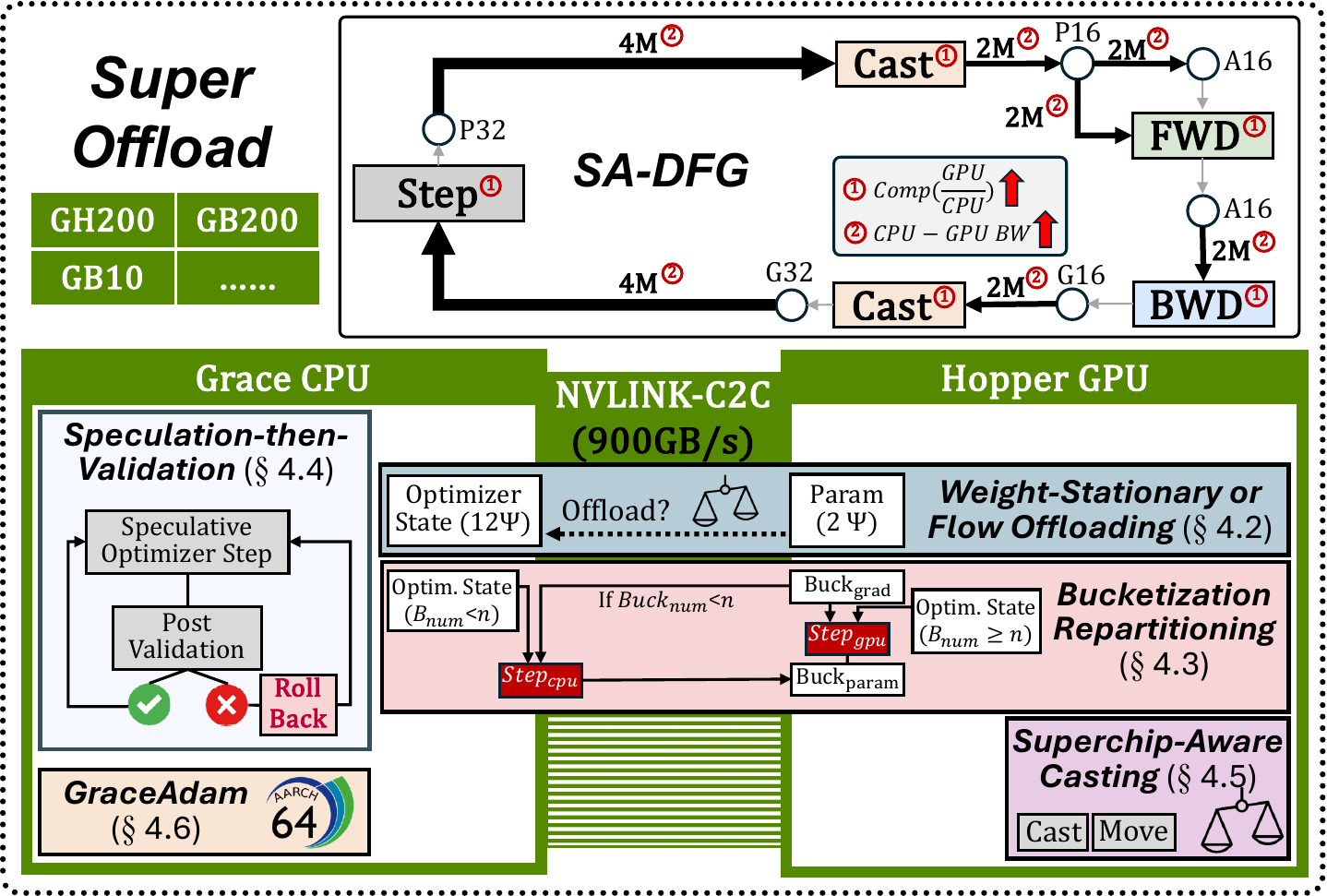}
    \caption{Overview of \superoffload. At a high level, \superoffload employs a Superchip-centric design that judiciously optimizes data placement, computation, and tensor migration between the Hopper GPU and Grace CPU, while fully utilizing the NVLink-C2C via a combination of techniques.}
    \label{fig:overview}
\end{figure}

\fref{fig:overview} illustrates an overview of \superoffload, a system specifically designed for training LLMs on Superchips by judiciously optimizing data placement, computation, and tensor migration between Hopper GPU and Grace CPU. This is achieved through a combination of techniques. First, it examines the amount of computation and communication associated with the operators in \scdfg, and employs an adaptive weight-stationary and weight-flow policy to exploit heterogeneous compute efficiencies of Hopper GPU and Grace CPU (\sref{subsec:weight-placement}). Second, it adjusts the bucketization strategy and leverages fine-grained repartitioning to balance the workload among different resources on a Superchip (\sref{subsec:bucketization}). Third, it introduces a novel speculation-then-validation algorithm to replace the conventional synchronization-then-execution scheme that enables effective overlapping between Hopper GPU and Grace CPU (\sref{subsec:speculation-schedule}). Fourth, it performs Superchip-aware casting for mixed-precision training (\sref{subsec:casting-cost}). Fifth, we develop a highly optimized Adam optimizer, GraceAdam, to make the best use of the Grace CPU  (\sref{subsec:graceadam}). Finally, \superoffload can be extended with ZeRO-style data parallelism and Ulysses-style sequence parallelism to support multi-Superchip training and training with long sequences (\sref{subsec:multi-superchip}).

\subsection{Adaptive Weight-Stationary and Weight-Flow Offloading}
\label{subsec:weight-placement}
When it comes to offloading LLM training, a key issue is to determine what model states to offload from GPU to CPU. While CPUs offer large memory capacity, transferring model states to and from CPU memory still incurs time delays. Given that both the forward propagation and backward propagation of LLM training have a computation complexity of $O(bsz\cdot\Psi)$ and the optimizer state update has a complexity of $O(\Psi)$, where $bsz$ is the batch size, state-of-the-art offloading systems often offload optimizer states and updates to the CPU~\cite{zero-offload,zero-infinity}, which significantly reduces GPU memory consumption while avoiding offloading compute-intensive operations to the CPU. 

However, an offloading decision must also be made for the model weights. Prior work often takes one of two directions: (1) \emph{weight-stationary}, which keeps model weights on GPU~\cite{zero-offload}, and (2) \emph{weight-flow}, which partially loads model weights part-by-part~\cite{zero-infinity}. For example, ZeRO-Offload keeps FP16 weights stationary on the GPU while offloading optimizer states to the CPU, reducing GPU memory consumption by roughly 8$\times$. Conversely, ZeRO-Infinity partially offloads FP16 weights to DRAM or NVMe, allowing larger models to be trained with a single GPU. Given the drastic change in Superchip architecture and emerging post-training scenarios like long-context extension, an important question arises: shall we offload the FP16 weights from Hopper GPU to the Grace CPU to scale up model training on \superchip?

To analyze the theoretical viability of offloading FP16 weights to Grace CPU on Superchip, 
we use the peak computational throughput ($peak_{tp}$), data movement bandwidth ($bw$) to estimate training efficiency.
The efficiency metric can be derived as follows:
{\small
\begin{align}
    comp\_time &= \frac{total\_computation}{peak_{tp}} \\
    comm\_time &= \frac{total\_data\_movement}{bw} \\
    efficiency &= \frac{comp\_time}{comp\_time + comm\_time}
\end{align}
}

For the forward pass in LLMs, the total computation is dominated by the operations in the linear layers. This can be approximated as a function of the number of parameters, sequence length, and batch size (i.e., $2 \times bsz \times seq \times params$). During the forward, the model parameters must be loaded from CPU to GPU at least once, resulting in $data\_movement$ of $2 \times params$ in bytes. To identify $peak_{tp}$, we use the achievable peak instead of the theoretical hardware peak. 

\begin{figure}[!t]
    \centering
    \includegraphics[width=\linewidth]{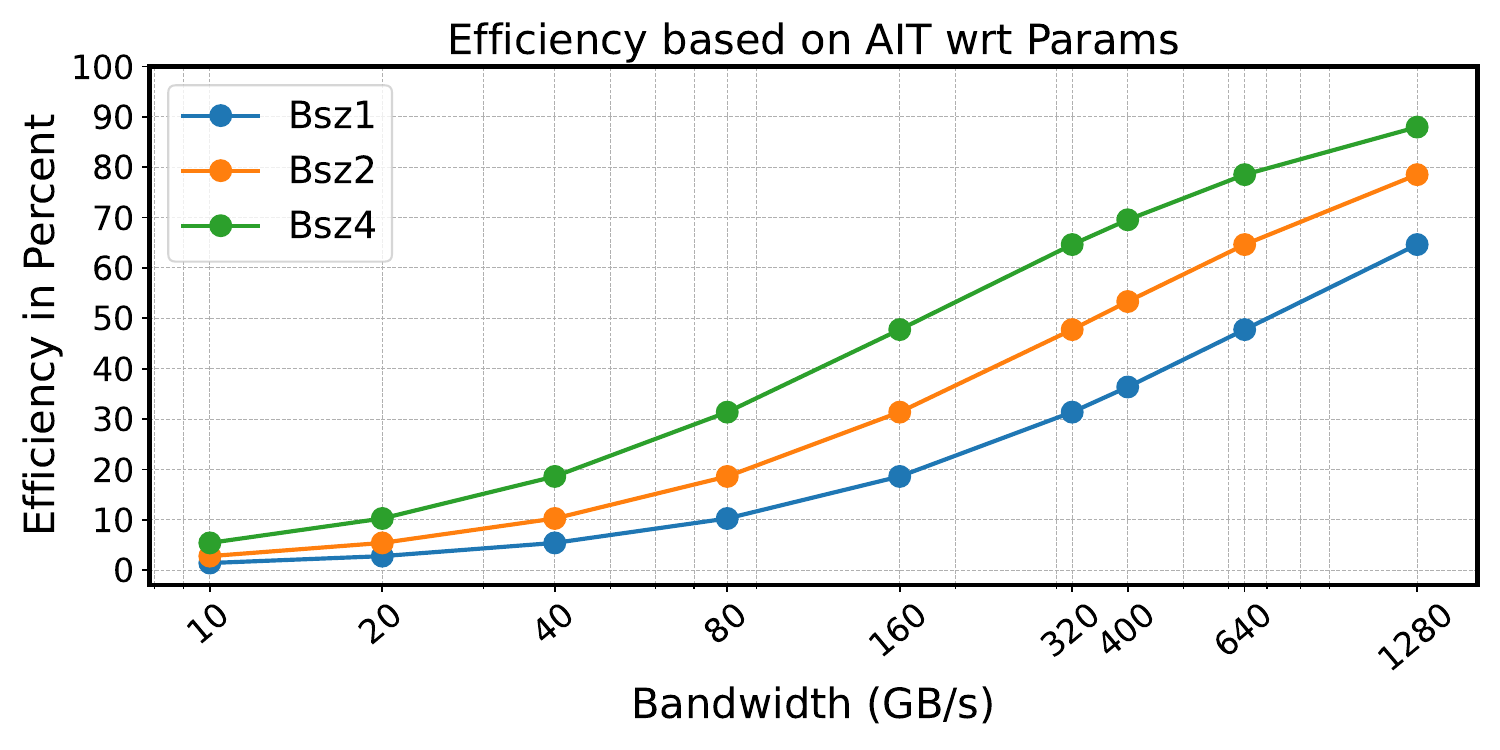}
    \caption{Impact of bandwidth on efficiency.}
    \label{fig:param_bw}
\end{figure}

To ensure that data movement can be completely overlapped with computation, the efficiency should exceed 50\% and ideally surpass 60\% when considering communication latency and other overhead. \fref{fig:param_bw} shows that even with a theoretical peak uni-directional C2C bandwidth of 450 GB/s, the batch size needs to be larger or equal to 4 with a sequence length of 1024 to achieve an efficiency greater than 60\%.

This analysis indicates that the optimal offloading strategy is scenario-dependent. When scaling model sizes, each GPU can only handle a small micro-batch size to prevent out-of-memory, as seen in training large-scale LLMs like  Bloom~\cite{bloom}, Megatron-LM~\cite{megatron-lm}, and Turing-NLG models~\cite{turing-nlg}, which often use a micro-batch size of 1 or 2. As batch size and sequence lengths grow, such as in post-training with long-context extension~\cite{longrope,longcontext}, 
activation memory may become significantly larger than model states. 
For instance, a 7B-parameter model requires 112GB for model states but needs 2TB of memory for activations with a sequence length of 1 million tokens.
When this happens, it becomes more economical to transfer FP16 weights between the Hopper GPU and Grace CPU. In \superoffload, we support both weight-stationary and weight-flow policies and adaptively choose the offloading policy for various practical settings.  
% For moderate batch sizes, it is beneficial to use a "hybrid" approach where both...  

\subsection{Fine-Grained Bucketization Repartitioning}
\label{subsec:bucketization}
One challenge in offloading-based solutions is that they cannot fully utilize both the computation and interconnect resources simultaneously because the GPU and CPU are idle during data movement operations. To overcome this challenge, prior work uses a \emph{bucketization} strategy~\cite{zero-offload}, dividing model states (e.g.,weights, gradients, optimizer states) into buckets to overlap GPU computation ( e.g., MatMul during backward propagation) with swap-out/in operations (e.g., transferring gradients from GPU to CPU) in a fine-grained manner. This allows the swap-out/in operation of a bucket to start as soon as the compute kernel produces it.

In theory, bucketization should allow GPU computation to fully overlap with the optimizer step on CPU and data transfer between GPU and CPU. However, in practice, we observe that Hopper GPU still experiences long idle times with bucketization, as shown in \fref{fig:idle}. 
By diving deeper into the issue, we find that this is primarily because the FLOPS ratio between Hopper GPU and Grace CPU is $\sim$330, which is significantly larger than prior architectures such as DGX-2 ($\sim$60.39) and DGX-A100 ($\sim$135.65). This increased compute gap makes balancing computation challenging. 
Take the last bucket as an example. In order for the last bucket's optimizer state to be executed on CPU, it requires a three-stage process: (i) gradient swap-out from GPU to CPU, (ii) execution of the Adam optimization algorithm on CPU, and (iii) swap-in the updated parameters back to the GPU. Since the first parameter in the forward pass is updated by the last gradient created in the backward pass, all of the above processes cannot overlap with any computation on GPU because the GPU needs to wait for the optimizer and weight updates to completely finish before starting the forward pass of the next training iteration. As a result, the latency from executing the last bucket(s) becomes exposed to the critical path length, and we need to carefully reorganize the computation to improve the Superchip utilization.

In contrast to previous studies~\cite{zero-offload}, \superoffload repartitions \scdfg by placing optimizer states and gradients for the last few buckets on the GPU, provided they fit into Hopper GPU memory. 
Moreover, to overlap the compute and communication, the last bucket of gradients offloaded to the CPU must complete its optimization step and return updated parameters to the GPU before the next iteration begins. 
If we denote the number of buckets with optimizer states kept on the GPU as $n$, then:
{\small
\begin{align}
    \operatorname{move_{bucket}}(grad) + \operatorname{step_{cpu}}(bucket) + \operatorname{move_{bucket}}(param) &\leq \\
    \operatorname{bwd}(n \cdot bucket) + \operatorname{step_{gpu}}(n \cdot bucket)
\end{align}
}
\noindent
$\operatorname{Step_{cpu/gpu}}(bucket)$ represents the time required to complete the optimizer step on the CPU/GPU for a bucket size of $bucket$.
$\operatorname{bwd}(n \cdot bucket)$ denotes the time to complete the backward pass for $n$ buckets on GPU, which can be approximated as a function of the number of parameters, sequence length, and batch size, given by $4 \times bsz \times seq \times bk \times n / 4$.

\begin{figure}[!t]
    \centering
    \includegraphics[width=\linewidth]{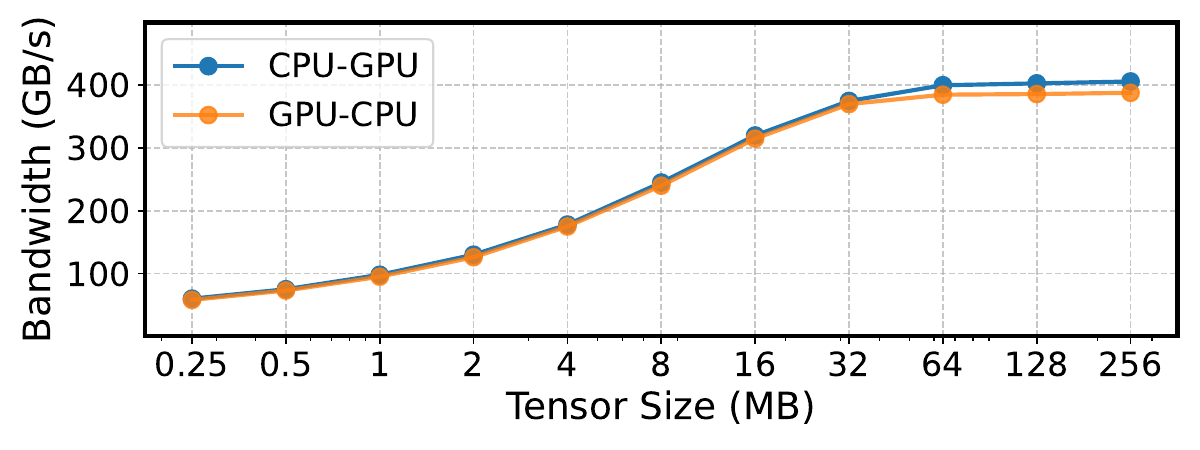}
    \caption{GH200 bandwidth measurement.}
    \vspace{-10pt}
    \label{fig:gpu-cpu_bw}
\end{figure}

The analysis indicates that the bucket size $bk$ is a crucial system parameter. \fref{fig:gpu-cpu_bw} shows the bandwidth between GPU and CPU across different tensor sizes. It shows that the C2C bandwidth increases with tensor size until saturation occurs at approximately 64 MB. Based on this observation, we implement a bucketization strategy where gradients and parameters are grouped into buckets of 64 MB before transfer between GPU and CPU. This approach ensures efficient utilization of the available C2C bandwidth while also providing a convenient unit that enables fine-grained overlapping between compute and communication. Finally, \superoffload determines the number of buckets to retain on GPU to hide Grace CPU execution and data transfer time. The optimal number of buckets depends on model size and batch sizes, and \superoffload uses grid search to identify the optimal number.

% , as the number of buckets for modern LLMs is often less than \textcolor{red}{XX}.   

\subsection{Speculation-then-Validation (STV)}
\label{subsec:speculation-schedule}

% As discussed earlier, the optimizer step on the CPU must wait until all gradients have been received. A key reason for this waiting period is that the optimizer's parameter update typically involves global operations—such as gradient clipping and normalization—that require a complete view of the gradients. As illustrated by the gray block in \fref{fig:zero-offload_schedule}, this dependency places the optimizer step for all parameters on the critical path, preventing it from overlapping with the backward pass.

In most offloading solutions, synchronizations between CPU and GPU are needed in optimizer step to ensure numerical robustness. As an example, the clipping of the gradient norm requires calculating the global gradient norm~\cite{gradient-norm}, and mixed precision training requires a global check of NAN and INF values~\cite{mixed-precision-training}. Both examples require the CPU to wait until all gradients have been received before the optimizer step and weight updates.
% transmitting updated parameters to GPU. 
As illustrated by the gray block in \fref{fig:zero-offload_schedule}, this dependency exposes the optimizer step to the critical path, preventing it from overlapping with the backward pass.

To address this limitation, we propose a \emph{speculation-then-validation} algorithm, which largely bypasses these synchronizations while preserving the exact convergence property of the training. Our mechanism is based on a key observation: \emph{most of the time the global states have no effects}. For example, gradient clipping is rarely triggered, especially after the initial warm-up phase when gradient variance significantly reduces~\cite{radam}. Similarly, mixed precision training rarely encounters NAN and INF, as a healthy training run 
should not have numerical instability issues.

\begin{figure}[!t]
    \centering
    \includegraphics[width=\linewidth]{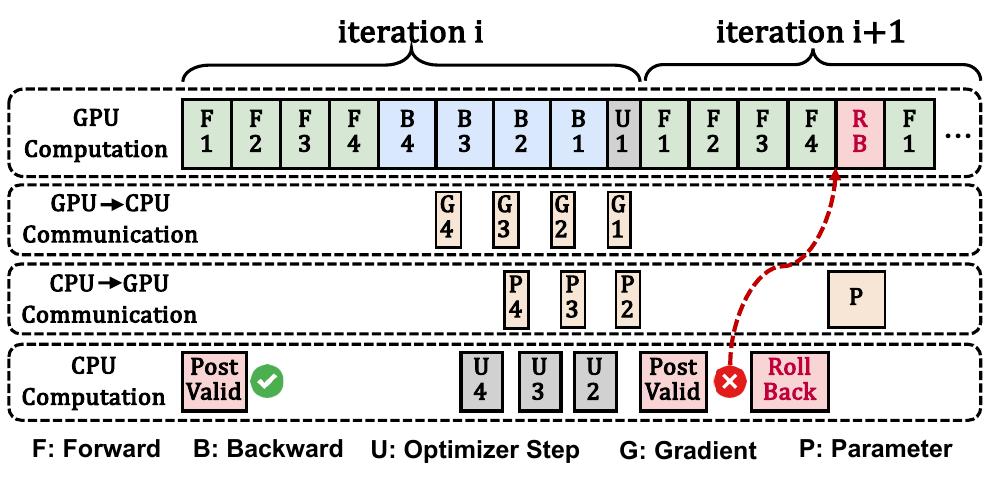}
    \caption{\superoffload speculation-then-validation schedule. It enables optimizer steps (U) to overlap with backward propagation (B), eliminating synchronization bottlenecks and improving resource utilization.}
    \label{fig:superoffload_schedule}
\end{figure}

Based on the observation, we propose to replace the conventional offloading synchronization-then-execution (STE) scheme with a speculation-then-validation (STV) mechanism. \fref{fig:superoffload_schedule} illustrates the idea. 
Instead of waiting for all gradients to arrive, the CPU initiates the optimizer step speculatively using the gradients available at that moment. Meanwhile, gradient clipping and normalization are deferred to occur during CPU idle periods, concurrent with the GPU executing the forward pass for the next iteration.
Meanwhile, to guarantee the synchronous nature of the optimization process, we implement the \emph{in-place rollback} for the optimization step. 

Therefore, STV is an exact optimization that preserves correctness by speculatively executing the optimizer step on CPU in parallel with the next GPU forward pass. A background process validates gradients in parallel by checking NaNs/Infs and gradient clipping violations. Specifically, the validation process is implemented using Python multiprocessing, and its results are passed to the GPU through a multiprocessing queue. After the forward pass, the GPU checks whether rollback is needed. There are two rollback scenarios: (1) if NaNs and Infs are detected, the iteration is skipped; (2) if gradients exceed clipping thresholds (e.g., after finishing computing the global gradient norm across all parameter gradients), SuperOffload reverts the previous optimizer update and re-executes it using the clipped gradients.

The speculation-then-validation scheduling offers two key benefits. First, it allows the optimizer steps on the CPU to overlap with backward propagation on the GPU, rather than placing all optimizer steps sequentially on the critical path, which would significantly hurt throughput. Second, it eliminates the time cost of normalization calculations and value checking by relocating these operations to previously idle CPU cycles (when the GPU is executing the forward pass), thereby improving overall resource utilization.
% Based on the multiple optimizations introduced in the previous subsection, the final schedule is shown in \fref{fig:superoffload_schedule}. 
Compared with the ZeRO-Offload schedule in \fref{fig:zero-offload_schedule}, \superoffload reduces the critical path of a single iteration from
{
\begin{align}
T_{\text{iter}} &= \operatorname{fwd_{gpu}} + \operatorname{bwd_{gpu}} + \operatorname{move_{bucket}}(grad) \nonumber\\
&\quad +\; \operatorname{step_{cpu}} + \operatorname{move_{bucket}}(param)
\end{align}
}
to
{
\begin{align}
T_{\text{iter}} = \operatorname{fwd_{gpu}} + \operatorname{bwd_{gpu}} + \operatorname{step_{cpu}}(bucket)
\end{align}
}

Interestingly, in theory, the STV schedule indicates that the iteration time of \superoffload can be shorter than that of non-offload training because the optimizer steps are performed on the entire model using GPU in non-offload training, while \superoffload only updates a subset of the parameters on GPU. However, the conventional wisdom is that offloading often comes at a performance cost. In our evaluation, we show that \superoffload can indeed be faster than GPU-only solutions. 
% Moreover, \superoffload enables training models that are up to 7$\times$ larger compared to those feasible with standard PyTorch training.

\subsection{Superchip-Aware Casting (SAC)}
\label{subsec:casting-cost}

In DL training frameworks like PyTorch and DeepSpeed, the mixed precision training is implemented through a graph rewriting process~\cite{amp}. The default precision of all ops is float32 (FP32). Mixed precision training casts certain model states (e.g., weights, gradients) from FP32 to float16 (FP16), or vice versa. For example, the gradients in the backward pass are produced in FP16, and the optimizer computes the updates using FP32 gradients. Therefore, the functions \(\operatorname{move_{bucket}}(grad)\) and \(\operatorname{move_{bucket}}(param)\) not only transfer tensors between the GPU and CPU but also involve converting tensor data types.
These additional casting operations become complex when considering offloading strategies. Existing offloading-based solutions often adopt a minimum edge cut algorithm to computation graph~\cite{zero-offload}, based on the implicit assumption that minimizing the data communication volume between the CPU and GPU leads to performance improvements.
However, these methods ignore the casting cost. Given that the C2C link between Hopper and Grace is an order of magnitude higher, is this assumption still valid? 
% Our performance modeling analysis shows that the cost associated with tensor casting is heavily affected by both casting cost and the data loading and saving times from or to external memory (see detailed analysis in Appendix~\ref{appendix:cast-analysis}). 

\begin{figure}[!t]
    \centering
    \includegraphics[width=0.95\linewidth]{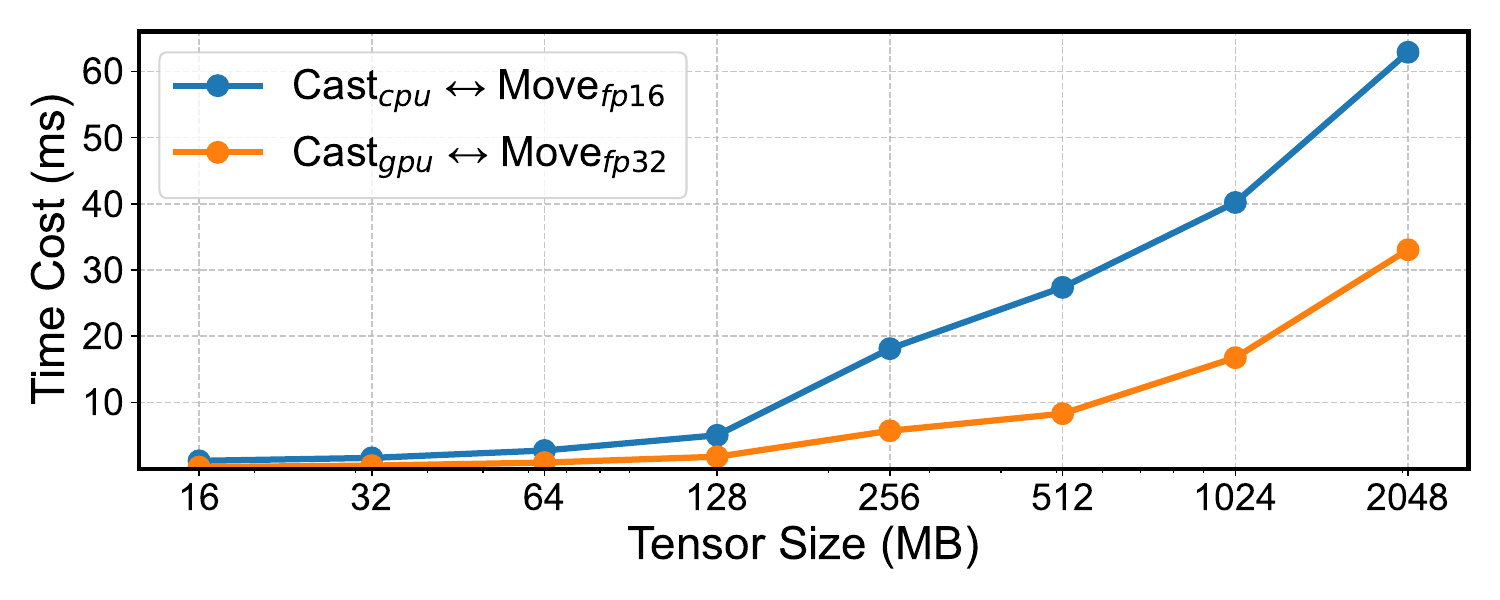}
    \caption{Time cost comparison for casting operations on GPU vs. CPU (including data transfer overhead).}
    \label{fig:cast_compare}
\end{figure}

We conduct experiments to examine the time cost of the two methods. The results are shown in \fref{fig:cast_compare}. We observe that $Cast_{cpu}\leftrightarrow\!Move_{fp16}$ takes around 2$\times$ execution time than $Cast_{gpu}\leftrightarrow\!Move_{fp32}$ on Superchips, depending on the size of the input tensor involved in mixed precision training. In most cases, $Cast_{gpu}\leftrightarrow\!Move_{fp32}$ surpasses the performance of $Cast_{cpu}\leftrightarrow\!Move_{fp16}$ by a large margin (e.g., tensor size 256MB to 2048MB). Based on our analysis, we opt for $Cast_{gpu}\leftrightarrow\!Move_{fp32}$, as it incurs an overall lower cost compared to $Cast_{cpu}\leftrightarrow\!Move_{fp16}$ even though the latter design incurs a lower theoretical communication volume.

In addition to the casting cost, we also find that the casting operation has an intricate interaction with the pinned memory. A transfer-then-cast operation triggers an unpinned temporary memory buffer allocation on the Grace CPU to hold the FP16 copy of gradients swapped-out from the Hopper GPU before casting FP16 gradients to FP32 on the Grace CPU. As such, the data transfer is implicitly through unpinned memory, which is significantly slower than DMA transfer with FP32. This observation also motivates us to adopt the $Cast_{gpu}\leftrightarrow\!Move_{fp32}$ design in \superoffload.

\subsection{GraceAdam}
\label{subsec:graceadam}

The careful reader may ask: Is Grace CPU in a Superchip sufficient for optimizer step and weight updates given the increased GPU/CPU FLOPS ratio? Prior work introduces CPU-Adam~\cite{zero-offload}, a highly-optimized implementation to accelerate Adam optimizer on CPU through multi-level parallelism.
% , which allows CPU-Adam to approach GPU-level performance on CPU.

However, the original CPU-Adam implementation was primarily designed for x86 architectures using AVX2/AVX512 instruction sets, which are not available on ARM CPUs. Moreover, while x86 implementations rely on fixed-width SIMD vectors, ARM's Scalable Vector Extension (SVE) employs a length-agnostic vector processing paradigm that requires more adaptive algorithms. Finally, ARM CPU has a different memory hierarchy, with different cache sizes, latencies, and bandwidth characteristics, which requires careful consideration of memory access patterns and prefetching strategies. 

To address this limitation, we introduce GraceAdam, which includes multiple optimizations: 1) \textbf{SVE Integration:} We adopt SVE's length-agnostic programming model to dynamically determine vector length at runtime with \texttt{svcntw()}. We replace scalar operations with SVE-specific intrinsics such as \texttt{svld1\_f32}, \texttt{svmla\_f32\_m}, and \texttt{svsqrt\_f32\_m} that operate on entire vectors simultaneously and thus reduce the computation cost. 2) \textbf{Enhanced Memory Management:}  We implement ARM-specific memory access patterns with explicit prefetching strategies using \texttt{svprfm} directives that preload data into cache before computation. Our tiled processing approach divides parameter updates into cache-friendly chunks (TILE size), minimizing data movement between memory hierarchies. We further optimize memory access by aligning SVE load/store operations (\texttt{svld1\_f32}, \texttt{svst1\_f32}) with ARM's cache line boundaries, reducing cache thrashing during parameter updates. 3) \textbf{Parallel Execution:} We employ a dual-level parallelism strategy, where we use OpenMP multithreading to ensure efficient scaling across multi-core ARM Grace CPUs and SVE for instruction-level parallelism within each core. These strategies together allow GraceAdam to maximize Grace CPU throughput and achieve significantly faster speed than the default PyTorch Adam on Grace CPU.  

\subsection{Multi-Superchip Schedule}
\label{subsec:multi-superchip}

In this part, we describe how \superoffload can be extended to work effectively in a multi-Superchip environment by combining it with ZeRO-style data parallelism (ZeRO-DP)~\cite{zero} and Ulysses-style sequence parallelism (Ulysses-SP)~\cite{ulysses}.

\vspace{7pt}
\noindent
\textbf{ZeRO-3 Integration.}
\superoffload preserves the model state partitioning strategy of ZeRO stage-3 (Fully Sharded Data Parallelism). Each GPU offloads a portion of its gradients and optimizer states to CPU memory while the partitioned weights, as well as the last few buckets from adaptive offloading (\sref{subsec:weight-placement}) remain on the GPUs. The primary advantage of partitioning before offloading is that in a multi-Superchip system, each parallel process is responsible for updating only a subset of the model parameters. This design ensures that the total communication volume from all GPUs to the CPU remains constant, while the CPU resources are utilized in parallel to jointly compute a unified weight update.

% \minjia{TODO: The current writing assumes weight-stationary. Need to add how weight-flow is handled.}
% \minjia{Sort of addressed it when discussing Super-Ulysses, because weight-flow is primarily used when training long sequences.}

% During forward propagation, the partitioned parameters are aggregated using a bucketized all-gather operation, similar to ZeRO-3.
% In the backward propagation phase, gradients are computed and averaged via a reduce-scatter operation on the GPUs. Subsequently, each GPU offloads only the averaged gradients corresponding to its partition -- and, for selected layers, depending on the offloading ratio (\sref{subsec:weight-placement}) -- to CPU memory. Once the CPU has received a bucket of gradients, the optimizer states are updated concurrently by each data-parallel process. Following the update, the partitioned parameters are transferred back to the GPUs.

\vspace{7pt}
\noindent
\textbf{\superulysses.}
\superoffload can also work together with Ulysses-SP. Recent advancements in large language models have demonstrated a clear trend toward training and fine-tuning with increasingly longer sequence lengths. Notable examples include Claude's expansion of context window from 100K to 200K tokens~\cite{claude} and GPT-4 Turbo's increase from 8K to 128K tokens~\cite{gpt4}. This evolution is driven by real-world applications that require processing extensive documents, images, and videos, requiring models capable of handling substantially longer sequences.

Ulysses-SP~\cite{ulysses} addresses sequence parallelism challenges by partitioning input data along the sequence dimension and employing an efficient all-to-all collective communication strategy for attention computations. 
However, even with activation checkpointing, the fixed GPU memory consumption of model states significantly limits the ability to scale to longer sequences. We integrate Ulysses-SP with \superoffload by leveraging its offloading capabilities to manage memory more efficiently. Specifically, \superoffload's adaptive weight-flow policy offloads optimizer states and the majority of model weights to CPU memory, significantly reducing the GPU memory footprint. This enables Ulysses-SP to handle longer sequences on a per-layer basis without being constrained by GPU memory limitations.
% since Ulysses only requires FP16 parameters that remain on the GPU side. 
% \minjia{The above statement is a bit confusing. The original Ulysses does not actually have this requirement of having only fp16 parameters on the GPU side. Maybe this was added by Ulysses-Offload? Also, based on Masahiro's suggestion, we should explain why SuperOffload works with Ulysses. @Xinyu, can you add a few sentences to explain how exactly \superoffload is integrated with Ulysses? Does that work out-of-the-box when you turn on Ulysses? }
% \xinyu{Yes, it works out-of-the-box, maybe we can say SuperOffload is a very general solutions, that can integrated with out code change with Ulysses, Universal checkpointing...}
% \minjia{I added some text. It does not say how this integration is done but at least provides some justification why Super-Ulysses provides a huge benefit -- the adaptive weight-flow policy in Superoffload offloads optimizer states and the majority of model weights to CPU. Please double check.}
We term the combined strategy \emph{\superulysses}. \superulysses supports 8$\times$ longer sequence training than Ulysses-SP and easily enables million-token sequence length LLM training with high throughput using only a few Superchips. See \sref{subsec:long-sequence-results} for more details. 

\vspace{7pt}
\noindent
\textbf{NUMA binding.}
In a K-way Superchip node, there are K Hopper GPUs and K Grace CPUs, and each Superchip is configured as a distinct NUMA node. Assume K = 4, a 4-way Superchip node has a total of 288 CPU cores (e.g., each Grace CPU has 72 cores). By default, a training launcher may randomly assign the process to certain CPU cores, which leads to issues such as the CPU process and GPU are not co-located on the same Superchip. If this happens, data transfers traverse an inter-Superchip connection (typically Slingshot), which has significantly lower bandwidth compared to the NVLINK-C2C connection. 
To mitigate this performance penalty, \superoffload explicitly binds each process to specific cores to ensure optimal CPU–GPU affinity.

\section{Evaluation}
\label{sec:eval}

In this section, we conduct a comprehensive evaluation of \superoffload in comparison with state-of-the-art approaches, demonstrating that it achieves excellent training efficiency and scalability across various model sizes and extremely long sequence training. We also show the impact of different technologies within \superoffload on overall performance.

% addressing the following research questions:
% \begin{enumerate}[label=(\roman*), leftmargin=*]
%     \item \textbf{Achievable Throughput:} How does the training throughput of \superoffload compare to state-of-the-art GPU-only and offloading-based solutions?
%     \item \textbf{Long Sequence Support:} To what extent does \superoffload enable training with extremely long sequences?
%     \item \textbf{Model Size Scalability:} How does \superoffload scale the trainable model size compared to existing multi-billion parameter training solutions?
% \end{enumerate}

\subsection{Evaluation Methodology}
\vspace{5pt}
\noindent
\textbf{Hardware.}
We conducted single \superchip experiments on a \textit{1$\times$GH200} (96GB HBM, 480GB DDR). For multiple nodes experiments, we used 8 \textit{GH200 NVL2} nodes, each containing \textit{2$\times$GH200} (96GB HBM, 240GB DDR), and the nodes are connected via HPE/Cray's 200Gbps Slingshot 11 interconnect.

\vspace{5pt}
\noindent
\textbf{Workloads.} 
For the performance evaluation, we focus on evaluating GPT/LLaMA~\cite{gpt-2,llama} like Transformer-based models~\cite{transformer}. 
We vary the hidden dimension and the number of Transformer blocks to obtain models with a different number of parameters. 
Appendix~\ref{sec:appendix:model-config} shows the configuration parameters used in our experiments. We use a subset of the Pile dataset~\cite{pile} for training datasets.

\vspace{5pt}
\noindent
\textbf{Baseline.} 
We compare the effectiveness of \superoffload with state-of-the-art multi-billion parameter training solutions: PyTorch DDP~\cite{ddp}, which uses data parallelism across GPUs; Megatron~\cite{megatron-lm}, which employs model parallelism; ZeRO-2/3~\cite{zero}, which shard gradients, optimizer states, and parameters (ZeRO-3) across GPUs; \zerooffload~\cite{zero-offload}, which extends ZeRO-2 by offloading optimzer states to CPU; ZeRO-Infinity~\cite{zero-infinity}, an extension of ZeRO-3 that offloads model states to CPU and NVMe (we only enable its CPU offloading for fair comparison); and FSDP-CPU Offload~\cite{fsdp}, which extends FSDP by offloading model states to CPU.  A more detailed description of baselines is provided in Appendix~\ref{sec:appendix:baseline}.

\subsection{Training Throughput}
\label{subsec:throughput}

\vspace{5pt}
\noindent
\textbf{Single \superchip.}
We compare the training throughput of \superoffload with PyTorch DDP, FSDP-Offload, ZeRO-Infinity and \zerooffload on a single \superchip across different model sizes. We exclude Megatron and ZeRO-2/3 from the comparison because they do not change the workload on a single GPU compared to PyTorch DDP, and none of them can train models exceeding 4B parameters due to out-of-memory (OOM) errors.  We evaluate them on the same batch size. And when large batch sizes cause OOM errors, we employ two strategies: (1) using smaller micro-batch sizes with gradient accumulation, or (2) implementing activation checkpointing with the largest micro-batch size without OOM errors. Note that we exclude recomputation volume when calculating TFLOPS to measure effective computational throughput. 
For each techniques, we report the higher throughput achieved between these two approaches.

\begin{figure}[!t]
    \centering
    \includegraphics[width=\linewidth]{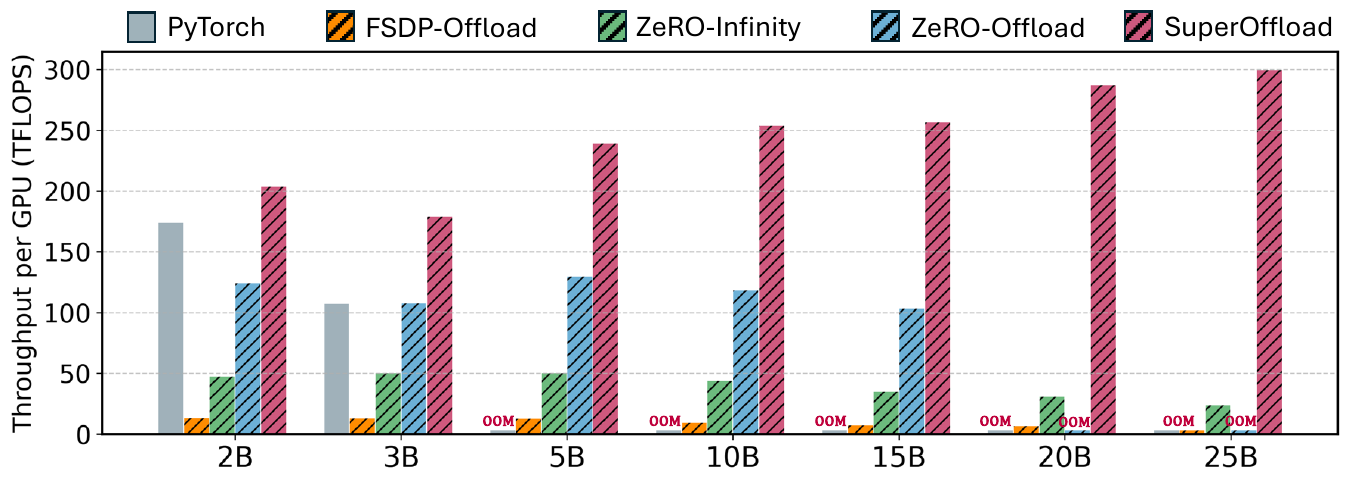}
    \caption{The training throughput with PyTorch DDP, FSDP-Offload, ZeRO-Infinity, \zerooffload, and \superoffload on a single \superchip with batch size 8.}
    \label{fig:eval_throughput_single_gpu}
\end{figure}

\fref{fig:eval_throughput_single_gpu} shows the comparison results. Interestingly, \superoffload not only outperforms existing offloading methods, but it also outperforms GPU-only approaches across all tested model sizes, which is different from conventional wisdom where offloading often comes at the cost of performance penalty. Specifically, \superoffload achieves up to 67\% higher throughput (TFLOPS) compared to PyTorch DDP. The reasons are two folds: (1) \superoffload offloads the optimizer steps to CPU, fully overlapping them with GPU backward propagation, and (2) by offloading the optimizer states to CPU memory, GPU has more memory space to hold the activations for larger batch without requiring activation checkpointing, which typically reduces throughput by approximately 33\%~\cite{tianqiac}. 
Moreover, \superoffload achieves 2$\times$ throughput on average (up to 2.5$\times$) compared to \zerooffload. The performance benefit of \superoffload comes from three aspects. First, the speculation-then-validation (\sref{subsec:speculation-schedule}) algorithm enables effective overlapping of CPU and GPU computation, whereas in \zerooffload, CPU optimizer steps must wait until GPU backward propagation completes. Second, the bucketization repartitioning (\sref{subsec:bucketization}) eliminates communication bottlenecks on the critical path when transferring gradients and updated parameters, 
while in \zerooffload, the forward pass for the next iteration must wait for updated parameters to be completely transferred from CPU to GPU. Third, GraceAdam (\sref{subsec:graceadam}) further reduces the computation cost of optimizer steps on Grace CPU.

\superoffload achieves significantly higher throughput than both FSDP-Offload and ZeRO-Infinity. FSDP-Offload consistently achieves less than 15 TFLOPS because it relies on PyTorch's native CPU-Adam implementation. 
As we demonstrate later in \sref{eval:adam}, native implementation is up to 3.5$\times$ slower than our GraceAdam, creating a substantial efficiency bottleneck. Similarly, ZeRO-Infinity's throughput remains below 50 TFLOPS. This limitation stems from ZeRO-Infinity's bucket-based approach for transferring gradients and parameters between CPU and GPU, which fails to account for the characteristics of C2C bandwidth—specifically, that bandwidth can drop to as low as 50GB/s with small tensor sizes.

\vspace{5pt}
\noindent
\textbf{Multiple Superchips.}
We evaluate the training throughput of \superoffload against Megatron, ZeRO-2, ZeRO-3, and ZeRO-Offload across 4 and 16 GH200 \superchips. We exclude the PyTorch DDP from the comparison as it cannot scale out to larger model sizes, and we exclude FSDP-offload and ZeRO-Infinity due to their significantly lower throughput compared to other solutions on single Superchip. 
% \minjia{@Xinyu, please address Masahiro's question by providing some numbers.}
% \xinyu{We have the number for the throuhgput on single superchip, for FSDP-Offload, consistently achieves less than 15 TFLOPS, remain below 50 TFLOPS.}
For Megatron, we use a MP degree that gives the best performance. Similar to the above single GPU experiment, when facing OOM errors, we either use gradient accumulation with smaller micro-batches or activation checkpointing with the largest possible micro-batch, reporting the higher throughput between these approaches.
\fref{fig:eval_throughput_node} shows the throughput per GPU results when training on 4 \superchips and 16 \superchips. We have the following observations:

\begin{figure}[!t]
    \centering
    \includegraphics[width=\linewidth]{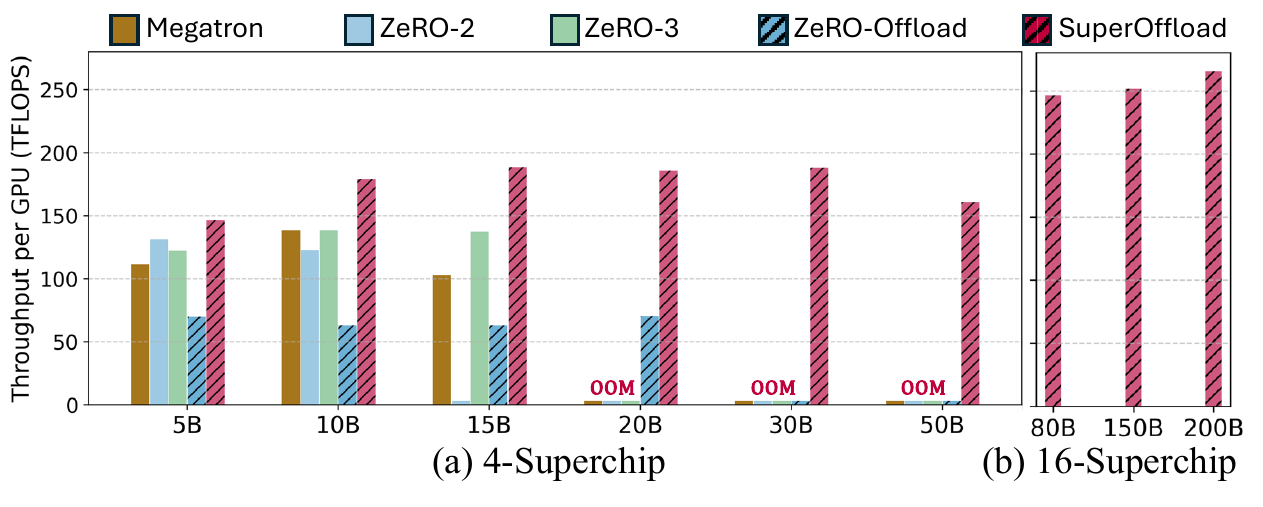}
    \caption{Training throughput comparison between Megatron, ZeRO-2/3, \zerooffload, and \superoffload on 4 and 16 GH200 GPUs. We use batch sizes of 16 and 128 for the 4-GPU and 16-GPU experiments, respectively.}
    \label{fig:eval_throughput_node}
\end{figure}

\begin{itemize}
    \item For models ranging from 5B to 20B parameters, \superoffload consistently achieves higher throughput compared to existing solutions. Specifically, it demonstrates throughput improvements of up to 83\%, 46\%, and 37\% over Megatron, ZeRO-2, ZeRO-3 respectively. In comparison with \zerooffload, \superoffload achieves 2.5$\times$ throughput on average across a wider range of model sizes. This is primarily attributed to highly overlapped CPU and GPU computation and sufficient GPU memory space for larger micro batch without using activation checkpointing, and those methods effectively leverage the high-bandwidth C2C interconnect in the \superchip architecture, enabling more efficient utilization of both GPU and CPU resources during training.
    \item While Megatron and ZeRO-3 encounter memory limitations at 15B parameters due to insufficient aggregate GPU memory across 4 GPUs, \superoffload effectively scales to models with up to 50B parameters.
    Moreover, \superoffload demonstrates exceptional scalability by efficiently training 200B models on 16 GPUs while maintaining high computational throughput.
\end{itemize}

\subsection{Unlock Massive Sequence Training}
\label{subsec:long-sequence-results}

\begin{figure}[!t]
    \centering
    \includegraphics[width=\linewidth]{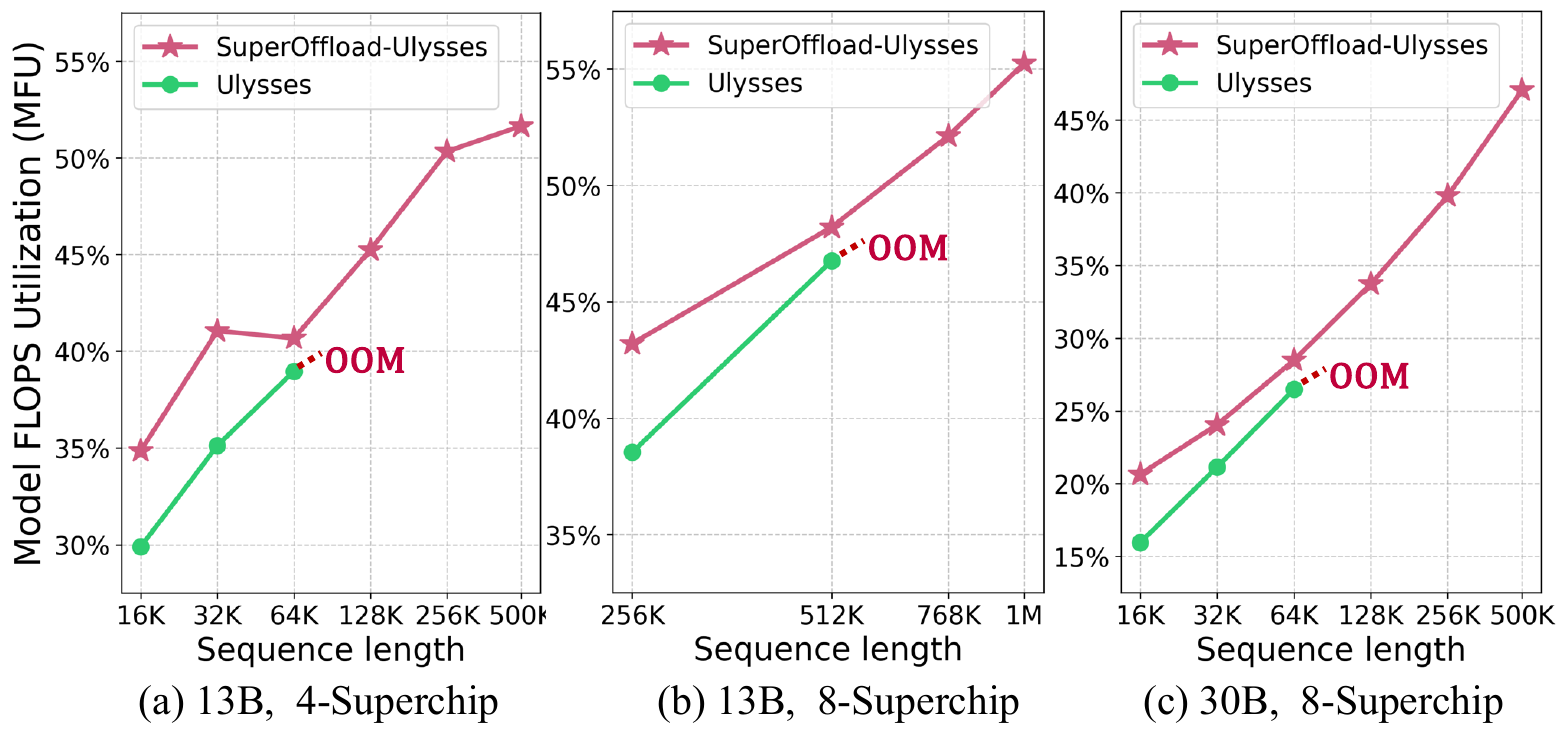}
    \caption{Supported sequence lengths and corresponding MFU using \superulysses and Ulysses. OOM denotes the point where increasing sequence length causes OOM.}
    \label{fig:eval_seq}
\end{figure}

We combine \superoffload together with Ulysses~\cite{ulysses} to create \superulysses for long-sequence training and evaluate its performance against vanilla Ulysses using two commonly deployed model sizes: 13B and 30B parameters. 
Our experiments are conducted on both 4-Superchip and 8-Superchip. For all experiments, we utilize the maximum batch sizes that avoid out-of-memory errors, and implemented activation checkpointing when activation sizes are substantial.
As shown in \fref{fig:eval_seq}, \superulysses trains sequences up to 8$\times$ longer than Ulysses. \superulysses enables the training of 13B model with sequence lengths up to 1 million tokens on 8 Superchips while achieving
55\% MFU. For sequence lengths that Ulysses can handle, \superulysses consistently achieves higher Model FLOPS Utilization (MFU). This performance advantage stems from our adaptive Offloading policy (\sref{subsec:weight-placement}), which adaptively detects extremely long sequences and dynamically offloads more weight parameters to CPU memory. 
This strategic offloading provides GPUs with additional memory space to accommodate larger activations without requiring activation checkpointing, thereby maintaining computational efficiency even with extended sequence lengths.

\subsection{Model Scale}
In this part, we first test the largest trainable models on a single \superchip as well as 4-\superchip and 16-\superchip.

\begin{figure}[!t]
    \centering
    \includegraphics[width=\linewidth]{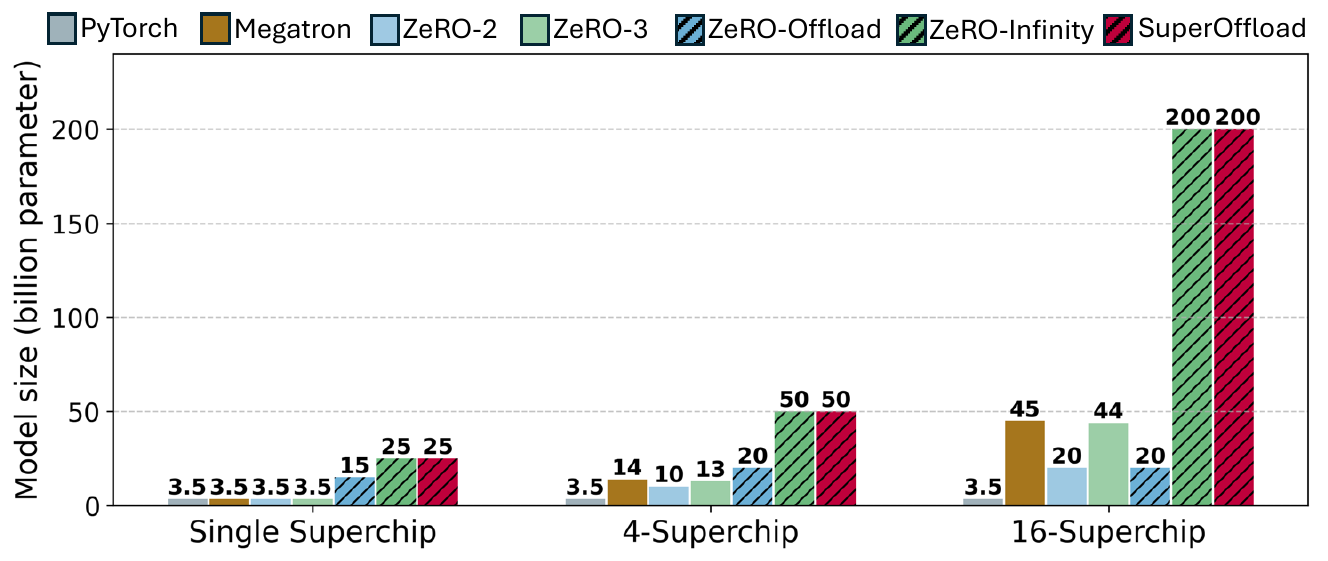}
    \caption{The size of the biggest model that can be trained on single \superchip, 4 and 16 \superchips.}
    \label{fig:eval_model_scale}
\end{figure}

\vspace{5pt}
\noindent
\textbf{Single Superchip.} 
\fref{fig:eval_model_scale} shows that
the largest model can be trained using PyTorch DDP on a single GPU with 96GB memory is 3.5B. Megatron, ZeRO-2, and ZeRO-3 do not enable training larger models on a single GPU compared to PyTorch DDP, because they all depend on the aggregated GPU memory to store model states. \zerooffload, however, can handle models up to 15B by offloading optimizer states. \superoffload can train models up to 25B -- primarily because it adaptively controls the offloading ratio for optimizer states and gradients. Unlike \zerooffload, which keeps weights stationary on GPUs and uses a fixed pattern for tensor placement across devices, \superoffload adaptively adopts weight-stationary and weight-flow policy based on the available GPU and CPU memory, thereby increasing the trainable model size.
While ZeRO-Infinity can train models of comparable size to \superoffload, its throughput is significantly lower. As shown in \fref{fig:eval_throughput_single_gpu}, \superoffload achieves on average 6.7$\times$ higher throughput (up to 12.6$\times$) than ZeRO-Infinity. This is mainly due to the technologies introduced by \superoffload such as STV and bucketization repartitioning that improve the overall utilization of the Hopper GPU and C2C bandwidth.
% \xinyu{This significant performance gap is primarily due to ZeRO-Infinity's use of very small bucket sizes (less than 1MB) that cannot be adjusted, resulting in suboptimal C2C bandwidth utilization as demonstrated in Figure 7. the other reasons including STV, repartiitoned that superoffload optimizerd with.}

\vspace{5pt}
\noindent
\textbf{Multiple Superchips.}
We further perform model scale tests with 4 and 16 GPUs in a single GH200 Node and four GH200 nodes, respectively. As shown in \fref{fig:eval_model_scale}, the maximum trainable model size stays the same for PyTorch DDP, 
because none of them handle memory redundancies in data parallelism. As a result, its scalability is bounded by the model scale on a single GPU. Megatron, ZeRO2/3 support large model training with more GPUs, but they cannot scale efficiently beyond 50B parameters, even with 16 GPUs. Megatron supports larger models than ZeRO-2, because ZeRO-2 still incurs memory redundancies on model weights. Since \zerooffload is built on top of ZeRO-2, it needs each GPU to hold at least the full copy of FP16 parameter, therefore, the largest model size the \zerooffload can train is bounded to 20B even though we increase the number of GPUs. Overall, \superoffload increases the model scale on 16 \superchips by 57$\times$, 4.4$\times$, 10$\times$, 4.5$\times$, 10$\times$ than using PyTorch DDP, Megatron, ZeRO-2, ZeRO-3, and \zerooffload, respectively.

\subsection{Performance Breakdown}
\label{subsec:breakdown}

We systematically enable each optimization component to quantify its individual contribution to overall performance. We use the same setting as used in \sref{subsec:throughput} and use 5B model as a demonstration.
\tref{tab:eval_breakdown} presents a comprehensive breakdown of \superoffload's key optimizations and their cumulative impact on training throughput. 
Starting with the baseline configuration (all optimizations disabled), we observe a throughput of 116.2 TFLOPS, that is close to the ZeRO-Offload throughput shown in \fref{fig:eval_throughput_single_gpu}. Enabling GraceAdam increases throughput to 128.23 TFLOPS, representing a 10.4\% improvement. When further enabling SAC from \sref{subsec:casting-cost}, throughput increases to 144.49 TFLOPS, delivering an additional 12.7\% improvement. This optimization strategically balances casting and data transfer costs for maximum efficiency. The Speculation-Then-Validation (STV) technique described in \sref{subsec:speculation-schedule} delivers the most substantial performance gain, boosting throughput by 45\%. 
This significant enhancement demonstrates how effectively STV eliminates Hopper GPU computational idle time by overlapping backward propagation with Grace CPU-based optimization steps. Finally, enabling the bucketization repartitioning from \sref{subsec:bucketization} further increases throughput to 238.92 TFLOPS, representing a 14.1\% improvement.
% \minjia{TODO: Xinyu, please double check the last sentence. Does "adaptive offloading mechanism" refer to both adaptive weight-stationary and weight-flow + bucketization repartitioning, or just one of them? }
% \xinyu{It just bucketization repartitioning.}
% \xinyu{The weight-stationary and weight-flow is only used for long-sequence training.}
% \minjia{Make sense. I updated the text. Comment out after you finished reading it.}
In total, \superoffload with all optimizations enabled achieves a 2.06$\times$ throughput improvement over the baseline configuration. This breakdown highlights how each component addresses specific performance bottlenecks in large-scale training on Superchip architectures, with their combined effect delivering substantial multiplicative benefits.

\begin{table}[!t]
\centering
\caption{Breakdown of \superoffload optimizations and their impact on throughput (TFLOPS), including GraceAdam (\sref{subsec:graceadam}), Superchip-Aware Casting (SAC, \sref{subsec:casting-cost}), Speculation-then-Validation (SVT, \sref{subsec:speculation-schedule}), and Bucket Repartitioning (\sref{subsec:bucketization}).}
{\small
\begin{tabular}{@{}c@{\hspace{4pt}}c@{\hspace{4pt}}c@{\hspace{5pt}}c@{\hspace{6pt}}|@{\hspace{4pt}}c@{}}
\toprule
\textbf{GraceAdam} & \textbf{Cast Optim.} & \textbf{STV} & \textbf{Buck. Repart.} & \textbf{Throughput} \\
\midrule
\xmark & \xmark & \xmark & \xmark & 116.20 \\
\cmark & \xmark & \xmark & \xmark & 128.23 \\
\cmark & \cmark & \xmark & \xmark & 144.49 \\
\cmark & \cmark & \cmark & \xmark & 209.36 \\
\cmark & \cmark & \cmark & \cmark & 238.92 \\
\bottomrule
\end{tabular}
\label{tab:eval_breakdown}
}
\end{table}

\subsection{Optimized GraceAdam Execution}
\label{eval:adam}

In this section, we evaluate our ARM-optimized GraceAdam implementation against the PyTorch CPU Adam and CPU-Adam implementation from \zerooffload (which was optimized for Intel CPUs).
\tref{tab:adam_latency} evaluates GraceAdam by measuring the optimizer execution times across model sizes ranging from 1B to 8B. 
% Our GraceAdam implementation leverages ARM's Scalable Vector Extension (SVE) for efficient SIMD parallelization, along with optimized memory management and OpenMP multithreading. 
Compared to PyTorch's CPU implementation (PT-CPU), GraceAdam achieves over 3$\times$ speedup across all configurations. Even compared to the already-optimized CPU-Adam implementation, GraceAdam demonstrates 1.36$\times$ faster execution.
These performance gains from GraceAdam's optimizations that leverage ARM's Scalable Vector Extension (SVE), along with optimized memory management and OpenMP multithreading, which make Grace CPU a viable and performant component within the \superoffload system.

% attributed to three key factors: (1) effective utilization of ARM SVE vector instructions, (2) cache-friendly memory access patterns that minimize data movement overhead, and (3) efficient parallel execution across multiple ARM cores. The results validate that our ARM-specific optimizations enable efficient parameter updates on Grace CPU, making Grace CPU a viable and performant component within the \superoffload system.

\begin{table}[!ht]
    \setlength{\tabcolsep}{8pt}
    \centering
    \caption{Adam latency (s) for PyTorch Native CPU-Adam, CPU-Adam, and  GraceAdam.}
    {\small
    \begin{tabular}{lccc}
        \toprule
        \textbf{\#Parameter} & \textbf{PT-CPU} & \textbf{CPU-Adam} & \textbf{GraceAdam} \\
        \midrule
        1 billion & 0.289  & 0.098  & 0.082  \\
        2 billion & 0.531  & 0.198  & 0.160  \\
        % 3 billion & 0.769  & 0.277  & 0.239  \\
        4 billion & 0.958  & 0.393  & 0.316  \\
        8 billion & 1.834  & 0.769  & 0.608  \\
        \bottomrule
    \end{tabular}
    }
    % \minjia{Can we go with the order PT-CPU, CPU-Adam, and GraceAdam? The convention of the order is (1) widely used baseline, (2) strong baseline, (3) our method. Update the caption if you change the order.}
    \label{tab:adam_latency}
\end{table}

\subsection{Speculation-Then-Validation Evaluation}
\label{subsec:eval:stv}

\begin{figure}[!b]
    \centering
    \includegraphics[width=0.95\linewidth]{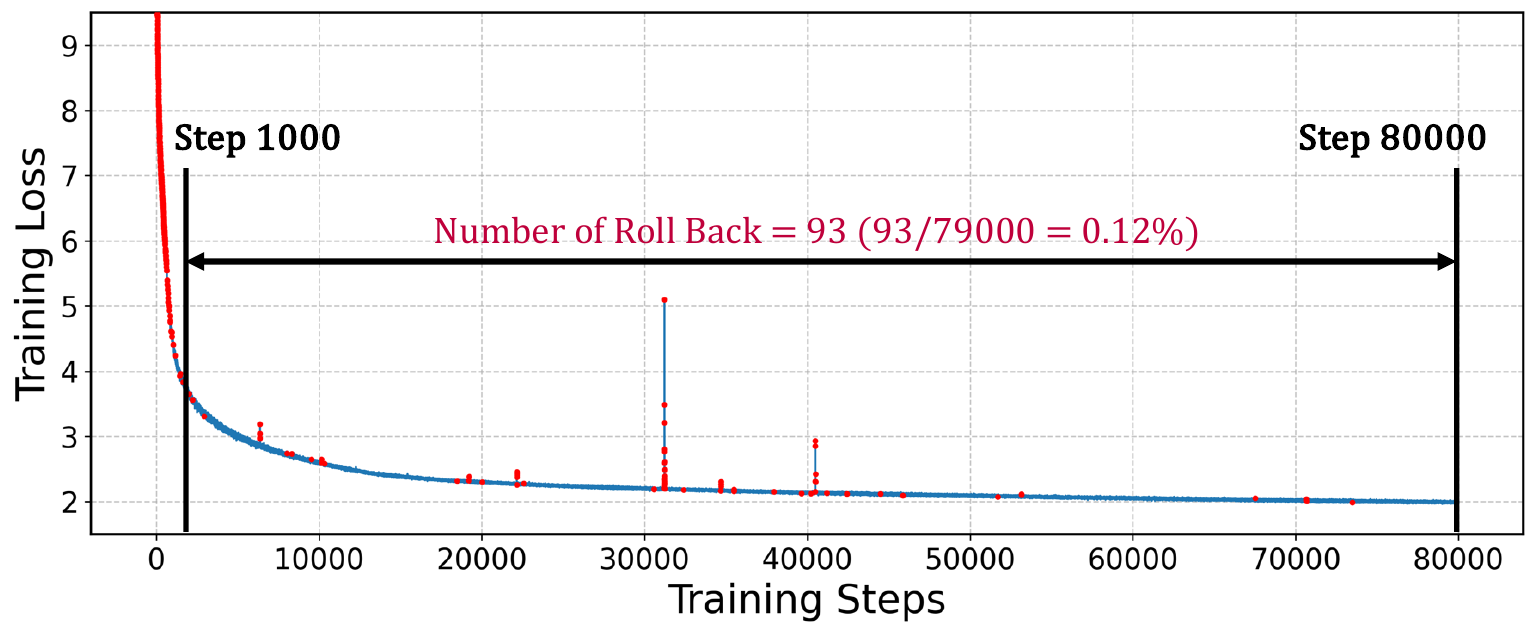}
    \caption{Training loss and rollback occurrences during training of the GPT 176B model over 80,000 iterations. Red dots indicate iterations where rollbacks were triggered due to gradient clipping, NaN or INF values.}
    % \minjia{"instances" should be "iterations"?}
    \label{fig:eval_roll-back}
\end{figure}

Speculation-then-Validation in \sref{subsec:speculation-schedule} is an exact optimization but the schedule is complex. To validate its correctness and overhead, we trained a GPT-style 175B model with the Pile dataset on 16 Superchips. Note that a GPT-style 175B model often requires thousands of GPUs to train~\cite{bloom,anyscale-175B,opt}, which are simply out of reach for most model scientists. 
Figure~\ref{fig:eval_roll-back} shows the pre-training loss curve over 80,000 training iterations, with the expected convergence trend. Though from iterations 1 to 1000, roll-backs occur frequently as gradient clipping and INF \& NaN values are mainly detected during the initial training stage. After iteration 1000, when training becomes more stable, roll-backs rarely happen - occurring only 93 times between steps 1000 and 80000, which represents 0.12\% of the total iterations. For the 175B model, roll-back operations executed in parallel across 16 Grace CPUs take 2 seconds on average, accumulating to less than 200 seconds over 79,000 iterations. This represents a negligible overhead while fully preserving training accuracy.

\subsection{GPU Idle Time Measurement}
\begin{figure}[!ht]
    \centering
    \includegraphics[width=\linewidth]{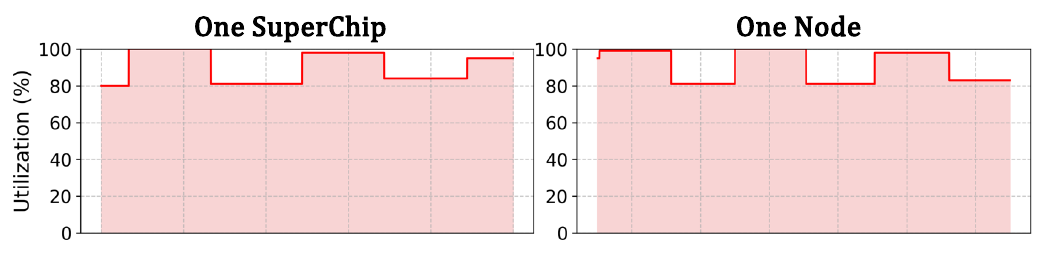}
    \caption{\superoffload fully utilizes the GPU resources.}
    \vspace{-8pt}
    \label{fig:superofflaod_idle}
\end{figure}

Unlike the prior solutions cause GPU idle time (as shown in \fref{fig:idle}), we measure the GPU utilization with the same model and hardware setting. As shown in \fref{fig:superofflaod_idle}, \superoffload achieves near-complete GPU utilization, effectively eliminating idle periods and maximizing computational efficiency.
\vspace{-12pt}

% \minjia{Do we have results that show the idle time, especially the GPU idle time, after applying optimizations in \superoffload? It would be a great call-back and also give the paper a strong ending if we have a figure similar to \fref{fig:idle} but with significantly reduced GPU idle time.}
% \xinyu{Yeah, I am adding this part.}

% \input{_s6_related}

\section{Conclusion}
\label{sec:conclusion}

In this paper, we have taken a leap forward in designing a Superchip-centric offloading system for large-scale LLM training. Our analysis reveals key challenges and performance implications of training LLMs on Superchips. Motivated by the observations, we design \superoffload, the first system that allows LLM training to make the best use of Hopper GPU, Grace CPU, and NVLink-C2C interconnects simultaneously. Our extensive evaluation demonstrates that \superoffload not only outperforms state-of-the-art offloading-based solutions but also unlocks new system capabilities such as training LLMs with 1 million token sequence length with only 8 Superchips, making large-scale LLM training more accessible to researchers and practitioners even without a large number of GPUs.

\section*{Acknowledgments}

We sincerely appreciate the anonymous reviewers and our shepherd Luo Mai. Their insightful feedback helps significantly improve the quality of the paper. We also gratefully acknowledge William Gropp, Brett Bode, and Gregory H. Bauer from the National Center for Supercomputing Applications (NCSA), as well as Dan Ernst, Ian Karlin, Giridhar Chukkapalli, Kurt Rago, and others from NVIDIA for their valuable discussions and guidance on MPAM support on Grace CPU. This research was supported by the National Science Foundation (NSF) under Grant No. 2441601. The work utilized the Delta and DeltaAI system at the National Center for Supercomputing Applications (NCSA) and Jetstream2 at Indiana University through allocation CIS240055 from the Advanced Cyberinfrastructure Coordination Ecosystem: Services \& Support (ACCESS) program, which is supported by National Science Foundation grants \#2138259, \#2138286, \#2138307, \#2137603, and \#2138296. The Delta advanced computing resource is a collaborative effort between the University of Illinois Urbana-Champaign and NCSA, supported by the NSF (award OAC 2005572) and the State of Illinois. UIUC SSAIL Lab is supported by research funding and gift from Google, IBM, and AMD. 

% This research was supported by the National Science Foundation (NSF) under Grant No. 2441601. The work utilized the DeltaAI system at the National Center for Supercomputing Applications (NCSA) through allocation CIS240055 from the Advanced Cyberinfrastructure Coordination Ecosystem: Services \& Support (ACCESS) program, which is supported by National Science Foundation grants \#2138259, \#2138286, \#2138307, \#2137603, and \#2138296. The Delta advanced computing resource is a collaborative effort between the University of Illinois Urbana-Champaign and NCSA, supported by the NSF (award OAC 2005572) and the State of Illinois. This work also utilized the Illinois Campus Cluster and NCSA NFI Hydro cluster, both supported by the University of Illinois Urbana-Champaign and the University of Illinois System.

\newpage
\clearpage
% use the ACM bibliography style
\balance
\bibliographystyle{ACM-Reference-Format}
\bibliography{reference}


\begin{thebibliography}{73}

%%% the Web address.

\ifx \showCODEN    \undefined \def \showCODEN     #1{\unskip}     \fi
\ifx \showISBNx    \undefined \def \showISBNx     #1{\unskip}     \fi
\ifx \showISBNxiii \undefined \def \showISBNxiii  #1{\unskip}     \fi
\ifx \showISSN     \undefined \def \showISSN      #1{\unskip}     \fi
\ifx \showLCCN     \undefined \def \showLCCN      #1{\unskip}     \fi
\ifx \shownote     \undefined \def \shownote      #1{#1}          \fi
\ifx \showarticletitle \undefined \def \showarticletitle #1{#1}   \fi
\ifx \showURL      \undefined \def \showURL       {\relax}        \fi
% invisible to TeX
\providecommand\bibfield[2]{#2}
\providecommand\bibinfo[2]{#2}
\providecommand\natexlab[1]{#1}
\providecommand\showeprint[2][]{arXiv:#2}

\bibitem[Abdin et~al\mbox{.}(2024)]%
        {phi3}
\bibfield{author}{\bibinfo{person}{Marah Abdin}, \bibinfo{person}{Jyoti Aneja}, \bibinfo{person}{Hany Awadalla}, \bibinfo{person}{Ahmed Awadallah}, \bibinfo{person}{Ammar~Ahmad Awan}, \bibinfo{person}{Nguyen Bach}, \bibinfo{person}{Amit Bahree}, \bibinfo{person}{Arash Bakhtiari}, \bibinfo{person}{Jianmin Bao}, \bibinfo{person}{Harkirat Behl}, \bibinfo{person}{Alon Benhaim}, \bibinfo{person}{Misha Bilenko}, \bibinfo{person}{Johan Bjorck}, \bibinfo{person}{Sébastien Bubeck}, \bibinfo{person}{Martin Cai}, \bibinfo{person}{Qin Cai}, \bibinfo{person}{Vishrav Chaudhary}, \bibinfo{person}{Dong Chen}, \bibinfo{person}{Dongdong Chen}, \bibinfo{person}{Weizhu Chen}, \bibinfo{person}{Yen-Chun Chen}, \bibinfo{person}{Yi-Ling Chen}, \bibinfo{person}{Hao Cheng}, \bibinfo{person}{Parul Chopra}, \bibinfo{person}{Xiyang Dai}, \bibinfo{person}{Matthew Dixon}, \bibinfo{person}{Ronen Eldan}, \bibinfo{person}{Victor Fragoso}, \bibinfo{person}{Jianfeng Gao}, \bibinfo{person}{Mei Gao}, \bibinfo{person}{Min Gao},
  \bibinfo{person}{Amit Garg}, \bibinfo{person}{Allie~Del Giorno}, \bibinfo{person}{Abhishek Goswami}, \bibinfo{person}{Suriya Gunasekar}, \bibinfo{person}{Emman Haider}, \bibinfo{person}{Junheng Hao}, \bibinfo{person}{Russell~J. Hewett}, \bibinfo{person}{Wenxiang Hu}, \bibinfo{person}{Jamie Huynh}, \bibinfo{person}{Dan Iter}, \bibinfo{person}{Sam~Ade Jacobs}, \bibinfo{person}{Mojan Javaheripi}, \bibinfo{person}{Xin Jin}, \bibinfo{person}{Nikos Karampatziakis}, \bibinfo{person}{Piero Kauffmann}, \bibinfo{person}{Mahoud Khademi}, \bibinfo{person}{Dongwoo Kim}, \bibinfo{person}{Young~Jin Kim}, \bibinfo{person}{Lev Kurilenko}, \bibinfo{person}{James~R. Lee}, \bibinfo{person}{Yin~Tat Lee}, \bibinfo{person}{Yuanzhi Li}, \bibinfo{person}{Yunsheng Li}, \bibinfo{person}{Chen Liang}, \bibinfo{person}{Lars Liden}, \bibinfo{person}{Xihui Lin}, \bibinfo{person}{Zeqi Lin}, \bibinfo{person}{Ce Liu}, \bibinfo{person}{Liyuan Liu}, \bibinfo{person}{Mengchen Liu}, \bibinfo{person}{Weishung Liu}, \bibinfo{person}{Xiaodong Liu},
  \bibinfo{person}{Chong Luo}, \bibinfo{person}{Piyush Madan}, \bibinfo{person}{Ali Mahmoudzadeh}, \bibinfo{person}{David Majercak}, \bibinfo{person}{Matt Mazzola}, \bibinfo{person}{Caio César~Teodoro Mendes}, \bibinfo{person}{Arindam Mitra}, \bibinfo{person}{Hardik Modi}, \bibinfo{person}{Anh Nguyen}, \bibinfo{person}{Brandon Norick}, \bibinfo{person}{Barun Patra}, \bibinfo{person}{Daniel Perez-Becker}, \bibinfo{person}{Thomas Portet}, \bibinfo{person}{Reid Pryzant}, \bibinfo{person}{Heyang Qin}, \bibinfo{person}{Marko Radmilac}, \bibinfo{person}{Liliang Ren}, \bibinfo{person}{Gustavo de Rosa}, \bibinfo{person}{Corby Rosset}, \bibinfo{person}{Sambudha Roy}, \bibinfo{person}{Olatunji Ruwase}, \bibinfo{person}{Olli Saarikivi}, \bibinfo{person}{Amin Saied}, \bibinfo{person}{Adil Salim}, \bibinfo{person}{Michael Santacroce}, \bibinfo{person}{Shital Shah}, \bibinfo{person}{Ning Shang}, \bibinfo{person}{Hiteshi Sharma}, \bibinfo{person}{Yelong Shen}, \bibinfo{person}{Swadheen Shukla}, \bibinfo{person}{Xia Song},
  \bibinfo{person}{Masahiro Tanaka}, \bibinfo{person}{Andrea Tupini}, \bibinfo{person}{Praneetha Vaddamanu}, \bibinfo{person}{Chunyu Wang}, \bibinfo{person}{Guanhua Wang}, \bibinfo{person}{Lijuan Wang}, \bibinfo{person}{Shuohang Wang}, \bibinfo{person}{Xin Wang}, \bibinfo{person}{Yu Wang}, \bibinfo{person}{Rachel Ward}, \bibinfo{person}{Wen Wen}, \bibinfo{person}{Philipp Witte}, \bibinfo{person}{Haiping Wu}, \bibinfo{person}{Xiaoxia Wu}, \bibinfo{person}{Michael Wyatt}, \bibinfo{person}{Bin Xiao}, \bibinfo{person}{Can Xu}, \bibinfo{person}{Jiahang Xu}, \bibinfo{person}{Weijian Xu}, \bibinfo{person}{Jilong Xue}, \bibinfo{person}{Sonali Yadav}, \bibinfo{person}{Fan Yang}, \bibinfo{person}{Jianwei Yang}, \bibinfo{person}{Yifan Yang}, \bibinfo{person}{Ziyi Yang}, \bibinfo{person}{Donghan Yu}, \bibinfo{person}{Lu Yuan}, \bibinfo{person}{Chenruidong Zhang}, \bibinfo{person}{Cyril Zhang}, \bibinfo{person}{Jianwen Zhang}, \bibinfo{person}{Li~Lyna Zhang}, \bibinfo{person}{Yi Zhang}, \bibinfo{person}{Yue Zhang},
  \bibinfo{person}{Yunan Zhang}, {and} \bibinfo{person}{Xiren Zhou}.} \bibinfo{year}{2024}\natexlab{}.
\newblock \showarticletitle{{Phi-3 Technical Report: A Highly Capable Language Model Locally on Your Phone}}.
\newblock \bibinfo{journal}{\emph{arXiv preprint arXiv:2404.14219}} (\bibinfo{year}{2024}).
\newblock


\bibitem[{AMD}(2024)]%
        {amd-mi300a}
\bibfield{author}{\bibinfo{person}{{AMD}}.} \bibinfo{year}{2024}\natexlab{}.
\newblock \bibinfo{title}{{AMD Instinct MI300A}}.
\newblock \bibinfo{howpublished}{\url{https://www.amd.com/en/products/accelerators/instinct/mi300/mi300a.html}}.
\newblock


\bibitem[{Anthropic}(2024)]%
        {claude}
\bibfield{author}{\bibinfo{person}{{Anthropic}}.} \bibinfo{year}{2024}\natexlab{}.
\newblock \bibinfo{title}{{Claude}}.
\newblock \bibinfo{howpublished}{\url{https://www.anthropic.com/claude}}.
\newblock


\bibitem[Anyscale(2023)]%
        {anyscale-175B}
\bibfield{author}{\bibinfo{person}{Anyscale}.} \bibinfo{year}{2023}\natexlab{}.
\newblock \bibinfo{title}{Training 175B Parameter Language Models at 1000 GPU scale with Alpa and Ray}.
\newblock \bibinfo{howpublished}{\url{https://www.anyscale.com/blog/training-175b-parameter-language-models-at-1000-gpu-scale-with-alpa-and-ray}}.
\newblock


\bibitem[Azure(2024)]%
        {azure-gh200}
\bibfield{author}{\bibinfo{person}{Microsoft Azure}.} \bibinfo{year}{2024}\natexlab{}.
\newblock \bibinfo{title}{Microsoft and NVIDIA partnership continues to deliver on the promise of AI}.
\newblock \bibinfo{howpublished}{\url{https://azure.microsoft.com/en-us/blog/microsoft-and-nvidia-partnership-continues-to-deliver-on-the-promise-of-ai}}.
\newblock


\bibitem[Brown et~al\mbox{.}(2020)]%
        {sft}
\bibfield{author}{\bibinfo{person}{Tom Brown}, \bibinfo{person}{Benjamin Mann}, \bibinfo{person}{Nick Ryder}, \bibinfo{person}{Melanie Subbiah}, \bibinfo{person}{Jared~D Kaplan}, \bibinfo{person}{Prafulla Dhariwal}, \bibinfo{person}{Arvind Neelakantan}, \bibinfo{person}{Pranav Shyam}, \bibinfo{person}{Girish Sastry}, \bibinfo{person}{Amanda Askell}, \bibinfo{person}{Sandhini Agarwal}, \bibinfo{person}{Ariel Herbert-Voss}, \bibinfo{person}{Gretchen Krueger}, \bibinfo{person}{Tom Henighan}, \bibinfo{person}{Rewon Child}, \bibinfo{person}{Aditya Ramesh}, \bibinfo{person}{Daniel Ziegler}, \bibinfo{person}{Jeffrey Wu}, \bibinfo{person}{Clemens Winter}, \bibinfo{person}{Chris Hesse}, \bibinfo{person}{Mark Chen}, \bibinfo{person}{Eric Sigler}, \bibinfo{person}{Mateusz Litwin}, \bibinfo{person}{Scott Gray}, \bibinfo{person}{Benjamin Chess}, \bibinfo{person}{Jack Clark}, \bibinfo{person}{Christopher Berner}, \bibinfo{person}{Sam McCandlish}, \bibinfo{person}{Alec Radford}, \bibinfo{person}{Ilya Sutskever}, {and}
  \bibinfo{person}{Dario Amodei}.} \bibinfo{year}{2020}\natexlab{}.
\newblock \showarticletitle{{Language Models are Few-Shot Learners}}. In \bibinfo{booktitle}{\emph{Proceedings of the 34th International Conference on Neural Information Processing Systems (NIPS'20)}}.
\newblock


\bibitem[Chen et~al\mbox{.}(2016)]%
        {tianqiac}
\bibfield{author}{\bibinfo{person}{Tianqi Chen}, \bibinfo{person}{Bing Xu}, \bibinfo{person}{Chiyuan Zhang}, {and} \bibinfo{person}{Carlos Guestrin}.} \bibinfo{year}{2016}\natexlab{}.
\newblock \showarticletitle{Training Deep Nets with Sublinear Memory Cost}.
\newblock \bibinfo{journal}{\emph{arXiv preprint arXiv:1604.06174}} (\bibinfo{year}{2016}).
\newblock


\bibitem[Chen et~al\mbox{.}(2024)]%
        {chen2024longlora}
\bibfield{author}{\bibinfo{person}{Yukang Chen}, \bibinfo{person}{Shengju Qian}, \bibinfo{person}{Haotian Tang}, \bibinfo{person}{Xin Lai}, \bibinfo{person}{Zhijian Liu}, \bibinfo{person}{Song Han}, {and} \bibinfo{person}{Jiaya Jia}.} \bibinfo{year}{2024}\natexlab{}.
\newblock \showarticletitle{LongLoRA: Efficient Fine-tuning of Long-Context Large Language Models}.
\newblock \bibinfo{journal}{\emph{arXiv preprint arXiv:2309.12307}} (\bibinfo{year}{2024}).
\newblock


\bibitem[Cui et~al\mbox{.}(2025)]%
        {cui2025delta}
\bibfield{author}{\bibinfo{person}{Shengkun Cui}, \bibinfo{person}{Archit Patke}, \bibinfo{person}{Hung Nguyen}, \bibinfo{person}{Aditya Ranjan}, \bibinfo{person}{Ziheng Chen}, \bibinfo{person}{Phuong Cao}, \bibinfo{person}{Brett Bode}, \bibinfo{person}{Gregory Bauer}, \bibinfo{person}{Catello~Di Martino}, \bibinfo{person}{Saurabh Jha}, \bibinfo{person}{Chandra Narayanaswami}, \bibinfo{person}{Daby Sow}, \bibinfo{person}{Zbigniew~T. Kalbarczyk}, {and} \bibinfo{person}{Ravishankar~K. Iyer}.} \bibinfo{year}{2025}\natexlab{}.
\newblock \showarticletitle{Characterizing GPU Resilience and Impact on AI/HPC Systems}.
\newblock \bibinfo{journal}{\emph{arXiv preprint arXiv:2503.11901}} (\bibinfo{year}{2025}).
\newblock


\bibitem[Ding et~al\mbox{.}(2024)]%
        {longrope}
\bibfield{author}{\bibinfo{person}{Yiran Ding}, \bibinfo{person}{Li~Lyna Zhang}, \bibinfo{person}{Chengruidong Zhang}, \bibinfo{person}{Yuanyuan Xu}, \bibinfo{person}{Ning Shang}, \bibinfo{person}{Jiahang Xu}, \bibinfo{person}{Fan Yang}, {and} \bibinfo{person}{Mao Yang}.} \bibinfo{year}{2024}\natexlab{}.
\newblock \showarticletitle{LongRoPE: Extending LLM Context Window Beyond 2 Million Tokens}.
\newblock \bibinfo{journal}{\emph{arXiv preprint arXiv:2402.13753}} (\bibinfo{year}{2024}).
\newblock


\bibitem[Fusco et~al\mbox{.}(2024)]%
        {superchip-data-study}
\bibfield{author}{\bibinfo{person}{Luigi Fusco}, \bibinfo{person}{Mikhail Khalilov}, \bibinfo{person}{Marcin Chrapek}, \bibinfo{person}{Giridhar Chukkapalli}, \bibinfo{person}{Thomas Schulthess}, {and} \bibinfo{person}{Torsten Hoefler}.} \bibinfo{year}{2024}\natexlab{}.
\newblock \showarticletitle{{Understanding Data Movement in Tightly Coupled Heterogeneous Systems: A Case Study with the Grace Hopper Superchip}}.
\newblock \bibinfo{journal}{\emph{arXiv preprint arXiv:2408.11556}} (\bibinfo{year}{2024}).
\newblock


\bibitem[Gao et~al\mbox{.}(2020)]%
        {pile}
\bibfield{author}{\bibinfo{person}{Leo Gao}, \bibinfo{person}{Stella Biderman}, \bibinfo{person}{Sid Black}, \bibinfo{person}{Laurence Golding}, \bibinfo{person}{Travis Hoppe}, \bibinfo{person}{Charles Foster}, \bibinfo{person}{Jason Phang}, \bibinfo{person}{Horace He}, \bibinfo{person}{Anish Thite}, \bibinfo{person}{Noa Nabeshima}, \bibinfo{person}{Shawn Presser}, {and} \bibinfo{person}{Connor Leahy}.} \bibinfo{year}{2020}\natexlab{}.
\newblock \showarticletitle{{The Pile: An 800GB Dataset of Diverse Text for Language Modeling}}.
\newblock \bibinfo{journal}{\emph{arXiv preprint arXiv:2101.00027}} (\bibinfo{year}{2020}).
\newblock


\bibitem[Gao et~al\mbox{.}(2024)]%
        {longcontext}
\bibfield{author}{\bibinfo{person}{Tianyu Gao}, \bibinfo{person}{Alexander Wettig}, \bibinfo{person}{Howard Yen}, {and} \bibinfo{person}{Danqi Chen}.} \bibinfo{year}{2024}\natexlab{}.
\newblock \showarticletitle{How to Train Long-Context Language Models (Effectively)}.
\newblock \bibinfo{journal}{\emph{arXiv preprint arXiv:2410.02660}} (\bibinfo{year}{2024}).
\newblock


\bibitem[Gholami et~al\mbox{.}(2024)]%
        {memory-wall}
\bibfield{author}{\bibinfo{person}{Amir Gholami}, \bibinfo{person}{Zhewei Yao}, \bibinfo{person}{Sehoon Kim}, \bibinfo{person}{Coleman Hooper}, \bibinfo{person}{Michael~W. Mahoney}, {and} \bibinfo{person}{Kurt Keutzer}.} \bibinfo{year}{2024}\natexlab{}.
\newblock \showarticletitle{{AI and Memory Wall}}.
\newblock \bibinfo{journal}{\emph{arXiv preprint arXiv:2403.14123}} (\bibinfo{year}{2024}).
\newblock


\bibitem[Hildebrand et~al\mbox{.}(2020)]%
        {autotm}
\bibfield{author}{\bibinfo{person}{Mark Hildebrand}, \bibinfo{person}{Jawad Khan}, \bibinfo{person}{Sanjeev Trika}, \bibinfo{person}{Jason Lowe-Power}, {and} \bibinfo{person}{Venkatesh Akella}.} \bibinfo{year}{2020}\natexlab{}.
\newblock \showarticletitle{{AutoTM: Automatic Tensor Movement in Heterogeneous Memory Systems using Integer Linear Programming}}. In \bibinfo{booktitle}{\emph{Proceedings of the 25th International Conference on Architectural Support for Programming Languages and Operating Systems (ASPLOS'20)}}.
\newblock


\bibitem[{HPCwire}(2024)]%
        {helios}
\bibfield{author}{\bibinfo{person}{{HPCwire}}.} \bibinfo{year}{2024}\natexlab{}.
\newblock \bibinfo{title}{HPE's New Helios Supercomputer Elevates Polish Scientific Research at Cyfronet}.
\newblock \bibinfo{howpublished}{\url{https://www.hpcwire.com/off-the-wire/hpes-new-helios-supercomputer-elevates-polish-scientific-research-at-cyfronet}}.
\newblock


\bibitem[Huang et~al\mbox{.}(2020)]%
        {swapadvisor}
\bibfield{author}{\bibinfo{person}{Chien-Chin Huang}, \bibinfo{person}{Gu Jin}, {and} \bibinfo{person}{Jinyang Li}.} \bibinfo{year}{2020}\natexlab{}.
\newblock \showarticletitle{{SwapAdvisor: Pushing Deep Learning Beyond the GPU Memory Limit via Smart Swapping}}. In \bibinfo{booktitle}{\emph{Proceedings of the 25th International Conference on Architectural Support for Programming Languages and Operating Systems (ASPLOS'20)}}.
\newblock


\bibitem[Huang et~al\mbox{.}(2019)]%
        {gpipe}
\bibfield{author}{\bibinfo{person}{Yanping Huang}, \bibinfo{person}{Youlong Cheng}, \bibinfo{person}{Ankur Bapna}, \bibinfo{person}{Orhan Firat}, \bibinfo{person}{Dehao Chen}, \bibinfo{person}{Mia~Xu Chen}, \bibinfo{person}{HyoukJoong Lee}, {and} \bibinfo{person}{et al.}} \bibinfo{year}{2019}\natexlab{}.
\newblock \showarticletitle{{GPipe: Efficient Training of Giant Neural Networks using Pipeline Parallelism}}. In \bibinfo{booktitle}{\emph{Proceedings of the 33rd International Conference on Neural Information Processing Systems (NIPS'19)}}.
\newblock


\bibitem[Hurt et~al\mbox{.}(2024)]%
        {superchip-study}
\bibfield{author}{\bibinfo{person}{J.~Alex Hurt}, \bibinfo{person}{Grant~J. Scott}, \bibinfo{person}{Derek Weitzel}, {and} \bibinfo{person}{Huijun Zhu}.} \bibinfo{year}{2024}\natexlab{}.
\newblock \showarticletitle{{Adventures with Grace Hopper AI Super Chip and the National Research Platform}}.
\newblock \bibinfo{journal}{\emph{arXiv preprint arXiv:2410.16487}} (\bibinfo{year}{2024}).
\newblock


\bibitem[{Information Technology Center, The University of Tokyo}(2024)]%
        {miyabi}
\bibfield{author}{\bibinfo{person}{{Information Technology Center, The University of Tokyo}}.} \bibinfo{year}{2024}\natexlab{}.
\newblock \bibinfo{title}{Miyabi Supercomputer System}.
\newblock \bibinfo{howpublished}{\url{https://www.cc.u-tokyo.ac.jp/en/supercomputer/miyabi/system.php}}.
\newblock


\bibitem[insideHPC(2024)]%
        {exa1-he}
\bibfield{author}{\bibinfo{person}{insideHPC}.} \bibinfo{year}{2024}\natexlab{}.
\newblock \bibinfo{title}{Eviden Delivers 104 PFLOPS NVIDIA-Powered EXA1-HE Supercomputer for CEA}.
\newblock \bibinfo{howpublished}{\url{https://insidehpc.com/2024/04/eviden-delivers-104-pflops-nvidia-powered-exa1-he-supercomputer-for-cea}}.
\newblock


\bibitem[Jacobs et~al\mbox{.}(2023)]%
        {ulysses}
\bibfield{author}{\bibinfo{person}{Sam~Ade Jacobs}, \bibinfo{person}{Masahiro Tanaka}, \bibinfo{person}{Chengming Zhang}, \bibinfo{person}{Minjia Zhang}, \bibinfo{person}{Shuaiwen~Leon Song}, \bibinfo{person}{Samyam Rajbhandari}, {and} \bibinfo{person}{Yuxiong He}.} \bibinfo{year}{2023}\natexlab{}.
\newblock \showarticletitle{DeepSpeed Ulysses: System Optimizations for Enabling Training of Extreme Long Sequence Transformer Models}.
\newblock \bibinfo{journal}{\emph{arXiv preprint arXiv:2309.14509}} (\bibinfo{year}{2023}).
\newblock


\bibitem[{Jülich Supercomputing Centre}(2024)]%
        {jupiter}
\bibfield{author}{\bibinfo{person}{{Jülich Supercomputing Centre}}.} \bibinfo{year}{2024}\natexlab{}.
\newblock \bibinfo{title}{JUPITER Supercomputer}.
\newblock \bibinfo{howpublished}{\url{https://www.fz-juelich.de/en/ias/jsc/jupiter/tech}}.
\newblock


\bibitem[Kaplan et~al\mbox{.}(2020)]%
        {scaleing-law-nlp}
\bibfield{author}{\bibinfo{person}{Jared Kaplan}, \bibinfo{person}{Sam McCandlish}, \bibinfo{person}{Tom Henighan}, \bibinfo{person}{Tom~B. Brown}, \bibinfo{person}{Benjamin Chess}, \bibinfo{person}{Rewon Child}, \bibinfo{person}{Scott Gray}, \bibinfo{person}{Alec Radford}, \bibinfo{person}{Jeffrey Wu}, {and} \bibinfo{person}{Dario Amodei}.} \bibinfo{year}{2020}\natexlab{}.
\newblock \showarticletitle{{Scaling Laws for Neural Language Models}}.
\newblock \bibinfo{journal}{\emph{arXiv preprint arXiv:2001.08361}} (\bibinfo{year}{2020}).
\newblock


\bibitem[Kingma and Ba(2015)]%
        {adam}
\bibfield{author}{\bibinfo{person}{Diederik~P. Kingma} {and} \bibinfo{person}{Jimmy Ba}.} \bibinfo{year}{2015}\natexlab{}.
\newblock \showarticletitle{{Adam: A Method for Stochastic Optimization}}. In \bibinfo{booktitle}{\emph{Proceedings of the 3rd International Conference on Learning Representations (ICLR'15)}}.
\newblock


\bibitem[Lambda(2024)]%
        {lambda-gh200}
\bibfield{author}{\bibinfo{person}{Lambda}.} \bibinfo{year}{2024}\natexlab{}.
\newblock \bibinfo{title}{NVIDIA GH200 Grace Hopper Superchip}.
\newblock \bibinfo{howpublished}{\url{https://lambdalabs.com/nvidia-gh200}}.
\newblock


\bibitem[Li et~al\mbox{.}(2020)]%
        {ddp}
\bibfield{author}{\bibinfo{person}{Shen Li}, \bibinfo{person}{Yanli Zhao}, \bibinfo{person}{Rohan Varma}, \bibinfo{person}{Omkar Salpekar}, \bibinfo{person}{Pieter Noordhuis}, \bibinfo{person}{Teng Li}, \bibinfo{person}{Adam Paszke}, {and} \bibinfo{person}{et al.}} \bibinfo{year}{2020}\natexlab{}.
\newblock \showarticletitle{{PyTorch Distributed: Experiences on Accelerating Data Parallel Training}}.
\newblock \bibinfo{journal}{\emph{arXiv preprint arXiv:2006.15704}} (\bibinfo{year}{2020}).
\newblock


\bibitem[Lian et~al\mbox{.}(2025)]%
        {lian2025large}
\bibfield{author}{\bibinfo{person}{Xinyu Lian}, \bibinfo{person}{Yinfang Chen}, \bibinfo{person}{Runxiang Cheng}, \bibinfo{person}{Jie Huang}, \bibinfo{person}{Parth Thakkar}, \bibinfo{person}{Minjia Zhang}, {and} \bibinfo{person}{Tianyin Xu}.} \bibinfo{year}{2025}\natexlab{}.
\newblock \showarticletitle{{Large Language Models as Configuration Validators}}. In \bibinfo{booktitle}{\emph{Proceedings of the 47th IEEE/ACM International Conference on Software Engineering (ICSE'25)}}.
\newblock


\bibitem[Liu et~al\mbox{.}(2023)]%
        {ring-attention}
\bibfield{author}{\bibinfo{person}{Hao Liu}, \bibinfo{person}{Matei Zaharia}, {and} \bibinfo{person}{Pieter Abbeel}.} \bibinfo{year}{2023}\natexlab{}.
\newblock \showarticletitle{{Ring Attention with Blockwise Transformers for Near-Infinite Context}}.
\newblock \bibinfo{journal}{\emph{arXiv preprint arXiv:2310.01889}} (\bibinfo{year}{2023}).
\newblock


\bibitem[Liu et~al\mbox{.}(2020)]%
        {radam}
\bibfield{author}{\bibinfo{person}{Liyuan Liu}, \bibinfo{person}{Haoming Jiang}, \bibinfo{person}{Pengcheng He}, \bibinfo{person}{Weizhu Chen}, \bibinfo{person}{Xiaodong Liu}, \bibinfo{person}{Jianfeng Gao}, {and} \bibinfo{person}{Jiawei Han}.} \bibinfo{year}{2020}\natexlab{}.
\newblock \showarticletitle{On the Variance of the Adaptive Learning Rate and Beyond}. In \bibinfo{booktitle}{\emph{Proceedings of the 8th International Conference on Learning Representations (ICLR'20)}}.
\newblock


\bibitem[Loshchilov and Hutter(2019)]%
        {adamw}
\bibfield{author}{\bibinfo{person}{Ilya Loshchilov} {and} \bibinfo{person}{Frank Hutter}.} \bibinfo{year}{2019}\natexlab{}.
\newblock \showarticletitle{{Decoupled Weight Decay Regularization}}. In \bibinfo{booktitle}{\emph{Proceedings of the 7th International Conference on Learning Representations (ICLR'19)}}.
\newblock


\bibitem[Maurya et~al\mbox{.}(2024)]%
        {dos}
\bibfield{author}{\bibinfo{person}{Avinash Maurya}, \bibinfo{person}{Jie Ye}, \bibinfo{person}{M.~Mustafa Rafique}, \bibinfo{person}{Franck Cappello}, {and} \bibinfo{person}{Bogdan Nicolae}.} \bibinfo{year}{2024}\natexlab{}.
\newblock \showarticletitle{{Deep Optimizer States: Towards Scalable Training of Transformer Models using Interleaved Offloading}}. In \bibinfo{booktitle}{\emph{Proceedings of the 25th International Middleware Conference (Middleware’24)}}.
\newblock


\bibitem[{Meta AI LLaMA Team}(2024)]%
        {llama3TechReport}
\bibfield{author}{\bibinfo{person}{{Meta AI LLaMA Team}}.} \bibinfo{year}{2024}\natexlab{}.
\newblock \bibinfo{title}{{The Llama 3 Herd of Models}}.
\newblock \bibinfo{howpublished}{\url{https://ai.meta.com/research/publications/the-llama-3-herd-of-models}}.
\newblock


\bibitem[Micikevicius et~al\mbox{.}(2018)]%
        {mixed-precision-training}
\bibfield{author}{\bibinfo{person}{Paulius Micikevicius}, \bibinfo{person}{Sharan Narang}, \bibinfo{person}{Jonah Alben}, \bibinfo{person}{Gregory~F. Diamos}, \bibinfo{person}{Erich Elsen}, \bibinfo{person}{David Garc{\'{\i}}a}, \bibinfo{person}{Boris Ginsburg}, \bibinfo{person}{Michael Houston}, \bibinfo{person}{Oleksii Kuchaiev}, \bibinfo{person}{Ganesh Venkatesh}, {and} \bibinfo{person}{Hao Wu}.} \bibinfo{year}{2018}\natexlab{}.
\newblock \showarticletitle{{Mixed Precision Training}}. In \bibinfo{booktitle}{\emph{Proceedings of the 6th International Conference on Learning Representations (ICLR'18)}}.
\newblock


\bibitem[{Microsoft Research}(2020)]%
        {turing-nlg}
\bibfield{author}{\bibinfo{person}{{Microsoft Research}}.} \bibinfo{year}{2020}\natexlab{}.
\newblock \bibinfo{title}{{Turing-NLG: A 17-billion-parameter language model by Microsoft}}.
\newblock \bibinfo{howpublished}{\url{https://www.microsoft.com/en-us/research/blog/turing-nlg-a-17-billion-parameter-language-model-by-microsoft}}.
\newblock


\bibitem[Narayanan et~al\mbox{.}(2019)]%
        {1f1b}
\bibfield{author}{\bibinfo{person}{Deepak Narayanan}, \bibinfo{person}{Aaron Harlap}, \bibinfo{person}{Amar Phanishayee}, \bibinfo{person}{Vivek Seshadri}, \bibinfo{person}{Nikhil~R. Devanur}, \bibinfo{person}{Gregory~R. Ganger}, \bibinfo{person}{Phillip~B. Gibbons}, {and} \bibinfo{person}{Matei Zaharia}.} \bibinfo{year}{2019}\natexlab{}.
\newblock \showarticletitle{{PipeDream: Generalized Pipeline Parallelism for DNN Training}}. In \bibinfo{booktitle}{\emph{Proceedings of the 27th ACM Symposium on Operating Systems Principles (SOSP'19)}}.
\newblock


\bibitem[Narayanan et~al\mbox{.}(2021)]%
        {megatron-lm-v2}
\bibfield{author}{\bibinfo{person}{Deepak Narayanan}, \bibinfo{person}{Mohammad Shoeybi}, \bibinfo{person}{Jared Casper}, \bibinfo{person}{Patrick LeGresley}, \bibinfo{person}{Mostofa Patwary}, \bibinfo{person}{Vijay~Anand Korthikanti}, \bibinfo{person}{Dmitri Vainbrand}, \bibinfo{person}{Prethvi Kashinkunti}, \bibinfo{person}{Julie Bernauer}, \bibinfo{person}{Bryan Catanzaro}, \bibinfo{person}{Amar Phanishayee}, {and} \bibinfo{person}{Matei Zaharia}.} \bibinfo{year}{2021}\natexlab{}.
\newblock \showarticletitle{{Efficient Large-Scale Language Model Training on GPU Clusters Using Megatron-LM}}.
\newblock \bibinfo{journal}{\emph{arXiv preprint arXiv:2104.04473}} (\bibinfo{year}{2021}).
\newblock


\bibitem[{National Center for Supercomputing Applications}(2024)]%
        {ncsa-deltai}
\bibfield{author}{\bibinfo{person}{{National Center for Supercomputing Applications}}.} \bibinfo{year}{2024}\natexlab{}.
\newblock \bibinfo{title}{Delta-AI}.
\newblock \bibinfo{howpublished}{\url{https://delta.ncsa.illinois.edu/delta-delta-ai}}.
\newblock


\bibitem[NVIDIA(2022)]%
        {dgx-a100}
\bibfield{author}{\bibinfo{person}{NVIDIA}.} \bibinfo{year}{2022}\natexlab{}.
\newblock \bibinfo{title}{NVIDIA DGX A100: The Universal System for AI Infrastructure}.
\newblock \bibinfo{howpublished}{\url{https://images.nvidia.com/aem-dam/Solutions/Data-Center/nvidia-dgx-a100-datasheet.pdf}}.
\newblock


\bibitem[{NVIDIA}(2023)]%
        {grace-hopper-architecture}
\bibfield{author}{\bibinfo{person}{{NVIDIA}}.} \bibinfo{year}{2023}\natexlab{}.
\newblock \bibinfo{title}{{NVIDIA Grace Hopper Superchip Architecture In-Depth}}.
\newblock \bibinfo{howpublished}{\url{https://developer.nvidia.com/blog/nvidia-grace-hopper-superchip-architecture-in-depth}}.
\newblock


\bibitem[{NVIDIA}(2024a)]%
        {gb200}
\bibfield{author}{\bibinfo{person}{{NVIDIA}}.} \bibinfo{year}{2024}\natexlab{a}.
\newblock \bibinfo{title}{{NVIDIA GB200 NVL72}}.
\newblock \bibinfo{howpublished}{\url{https://www.nvidia.com/en-us/data-center/gb200-nvl72}}.
\newblock


\bibitem[{NVIDIA}(2024b)]%
        {grace-hopper}
\bibfield{author}{\bibinfo{person}{{NVIDIA}}.} \bibinfo{year}{2024}\natexlab{b}.
\newblock \bibinfo{title}{{NVIDIA Grace Hopper Superchip}}.
\newblock \bibinfo{howpublished}{\url{https://www.nvidia.com/en-us/data-center/grace-hopper-superchip}}.
\newblock


\bibitem[{OpenAI}(2023)]%
        {gpt4}
\bibfield{author}{\bibinfo{person}{{OpenAI}}.} \bibinfo{year}{2023}\natexlab{}.
\newblock \bibinfo{title}{{GPT-4}}.
\newblock \bibinfo{howpublished}{\url{https://openai.com/product/gpt-4}}.
\newblock


\bibitem[Ouyang et~al\mbox{.}(2022)]%
        {rlhf}
\bibfield{author}{\bibinfo{person}{Long Ouyang}, \bibinfo{person}{Jeff Wu}, \bibinfo{person}{Xu Jiang}, \bibinfo{person}{Diogo Almeida}, \bibinfo{person}{Carroll~L. Wainwright}, \bibinfo{person}{Pamela Mishkin}, \bibinfo{person}{Chong Zhang}, \bibinfo{person}{Sandhini Agarwal}, \bibinfo{person}{Katarina Slama}, \bibinfo{person}{Alex Ray}, \bibinfo{person}{John Schulman}, \bibinfo{person}{Jacob Hilton}, \bibinfo{person}{Fraser Kelton}, \bibinfo{person}{Luke Miller}, \bibinfo{person}{Maddie Simens}, \bibinfo{person}{Amanda Askell}, \bibinfo{person}{Peter Welinder}, \bibinfo{person}{Paul Christiano}, \bibinfo{person}{Jan Leike}, {and} \bibinfo{person}{Ryan Lowe}.} \bibinfo{year}{2022}\natexlab{}.
\newblock \showarticletitle{{Training Language Models to Follow Instructions with Human Feedback}}.
\newblock \bibinfo{journal}{\emph{arXiv preprint arXiv:2203.02155}} (\bibinfo{year}{2022}).
\newblock


\bibitem[Pascanu et~al\mbox{.}(2013)]%
        {gradient-norm}
\bibfield{author}{\bibinfo{person}{Razvan Pascanu}, \bibinfo{person}{Tom{\'{a}}s Mikolov}, {and} \bibinfo{person}{Yoshua Bengio}.} \bibinfo{year}{2013}\natexlab{}.
\newblock \showarticletitle{On the Difficulty of Training Recurrent Neural Networks}. In \bibinfo{booktitle}{\emph{Proceedings of the 30th International Conference on Machine Learning, (ICML'13)}}.
\newblock


\bibitem[Peng et~al\mbox{.}(2020)]%
        {capuchin}
\bibfield{author}{\bibinfo{person}{Xuan Peng}, \bibinfo{person}{Xuanhua Shi}, \bibinfo{person}{Hulin Dai}, \bibinfo{person}{Hai Jin}, \bibinfo{person}{Weiliang Ma}, \bibinfo{person}{Qian Xiong}, \bibinfo{person}{Fan Yang}, {and} \bibinfo{person}{Xuehai Qian}.} \bibinfo{year}{2020}\natexlab{}.
\newblock \showarticletitle{{Capuchin: Tensor-based GPU Memory Management for Deep Learning}}. In \bibinfo{booktitle}{\emph{Proceedings of the 25th International Conference on Architectural Support for Programming Languages and Operating Systems (ASPLOS'20)}}.
\newblock


\bibitem[{PyTorch Team}(2025a)]%
        {amp}
\bibfield{author}{\bibinfo{person}{{PyTorch Team}}.} \bibinfo{year}{2025}\natexlab{a}.
\newblock \bibinfo{title}{{Automatic Mixed Precision}}.
\newblock \bibinfo{howpublished}{\url{https://pytorch.org/tutorials/recipes/recipes/amp_recipe.html}}.
\newblock


\bibitem[{PyTorch Team}(2025b)]%
        {fsdp-offload}
\bibfield{author}{\bibinfo{person}{{PyTorch Team}}.} \bibinfo{year}{2025}\natexlab{b}.
\newblock \bibinfo{title}{{FSDP Tutorial}}.
\newblock \bibinfo{howpublished}{\url{https://pytorch.org/tutorials/intermediate/FSDP_tutorial.html}}.
\newblock


\bibitem[Radford et~al\mbox{.}(2019)]%
        {gpt-2}
\bibfield{author}{\bibinfo{person}{Alec Radford}, \bibinfo{person}{Jeff Wu}, \bibinfo{person}{Rewon Child}, \bibinfo{person}{David Luan}, \bibinfo{person}{Dario Amodei}, {and} \bibinfo{person}{Ilya Sutskever}.} \bibinfo{year}{2019}\natexlab{}.
\newblock \bibinfo{booktitle}{\emph{{Language Models are Unsupervised Multitask Learners}}}.
\newblock \bibinfo{type}{{T}echnical {R}eport}. \bibinfo{institution}{OpenAI}.
\newblock


\bibitem[Rafailov et~al\mbox{.}(2024)]%
        {dpo}
\bibfield{author}{\bibinfo{person}{Rafael Rafailov}, \bibinfo{person}{Archit Sharma}, \bibinfo{person}{Eric Mitchell}, \bibinfo{person}{Stefano Ermon}, \bibinfo{person}{Christopher~D. Manning}, {and} \bibinfo{person}{Chelsea Finn}.} \bibinfo{year}{2024}\natexlab{}.
\newblock \showarticletitle{{Direct Preference Optimization: Your Language Model is Secretly a Reward Model}}.
\newblock \bibinfo{journal}{\emph{arXiv preprint arXiv:2305.18290}} (\bibinfo{year}{2024}).
\newblock


\bibitem[Rajbhandari et~al\mbox{.}(2019)]%
        {zero}
\bibfield{author}{\bibinfo{person}{Samyam Rajbhandari}, \bibinfo{person}{Jeff Rasley}, \bibinfo{person}{Olatunji Ruwase}, {and} \bibinfo{person}{Yuxiong He}.} \bibinfo{year}{2019}\natexlab{}.
\newblock \showarticletitle{{ZeRO: Memory Optimization Towards Training {A} Trillion Parameter Models}}.
\newblock \bibinfo{journal}{\emph{arXiv preprint arXiv:1910.02054}} (\bibinfo{year}{2019}).
\newblock


\bibitem[Rajbhandari et~al\mbox{.}(2021)]%
        {zero-infinity}
\bibfield{author}{\bibinfo{person}{Samyam Rajbhandari}, \bibinfo{person}{Olatunji Ruwase}, \bibinfo{person}{Jeff Rasley}, \bibinfo{person}{Shaden Smith}, {and} \bibinfo{person}{Yuxiong He}.} \bibinfo{year}{2021}\natexlab{}.
\newblock \showarticletitle{{ZeRO-Infinity: Breaking the GPU Memory Wall for Extreme Scale Deep Learning}}.
\newblock \bibinfo{journal}{\emph{arXiv preprint arXiv:2104.07857}} (\bibinfo{year}{2021}).
\newblock


\bibitem[Rasley et~al\mbox{.}(2020)]%
        {deepspeed}
\bibfield{author}{\bibinfo{person}{Jeff Rasley}, \bibinfo{person}{Samyam Rajbhandari}, \bibinfo{person}{Olatunji Ruwase}, {and} \bibinfo{person}{Yuxiong He}.} \bibinfo{year}{2020}\natexlab{}.
\newblock \showarticletitle{{DeepSpeed: System Optimizations Enable Training Deep Learning Models with Over 100 Billion Parameters}}. In \bibinfo{booktitle}{\emph{Proceedings of the 26th {ACM} {SIGKDD} Conference on Knowledge Discovery and Data Mining (KDD'20)}}.
\newblock


\bibitem[Ren et~al\mbox{.}(2021a)]%
        {sentinel}
\bibfield{author}{\bibinfo{person}{Jie Ren}, \bibinfo{person}{Jiaolin Luo}, \bibinfo{person}{Kai Wu}, \bibinfo{person}{Minjia Zhang}, \bibinfo{person}{Hyeran Jeon}, {and} \bibinfo{person}{Dong Li}.} \bibinfo{year}{2021}\natexlab{a}.
\newblock \showarticletitle{{Sentinel: Efficient Tensor Migration and Allocation on Heterogeneous Memory Systems for Deep Learning}}. In \bibinfo{booktitle}{\emph{Proceedings of the 27th International Symposium on High-Performance Computer Architecture (HPCA'21)}}.
\newblock


\bibitem[Ren et~al\mbox{.}(2021b)]%
        {zero-offload}
\bibfield{author}{\bibinfo{person}{Jie Ren}, \bibinfo{person}{Samyam Rajbhandari}, \bibinfo{person}{Reza~Yazdani Aminabadi}, \bibinfo{person}{Olatunji Ruwase}, \bibinfo{person}{Shuangyan Yang}, \bibinfo{person}{Minjia Zhang}, \bibinfo{person}{Dong Li}, {and} \bibinfo{person}{Yuxiong He}.} \bibinfo{year}{2021}\natexlab{b}.
\newblock \showarticletitle{{ZeRO-Offload: Democratizing Billion-Scale Model Training}}. In \bibinfo{booktitle}{\emph{Proceedings of the 2021 USENIX Annual Technical Conference (ATC'21)}}.
\newblock


\bibitem[Rhu et~al\mbox{.}(2016)]%
        {vdnn}
\bibfield{author}{\bibinfo{person}{Minsoo Rhu}, \bibinfo{person}{Natalia Gimelshein}, \bibinfo{person}{Jason Clemons}, \bibinfo{person}{Arslan Zulfiqar}, {and} \bibinfo{person}{Stephen~W. Keckler}.} \bibinfo{year}{2016}\natexlab{}.
\newblock \showarticletitle{{vDNN: Virtualized Deep Neural Networks for Scalable, Memory-Efficient Neural Network Design}}. In \bibinfo{booktitle}{\emph{Proceedings of the 49th Annual IEEE/ACM International Symposium on Microarchitecture (MICRO-49)}}.
\newblock


\bibitem[Rozière et~al\mbox{.}(2024)]%
        {codellama}
\bibfield{author}{\bibinfo{person}{Baptiste Rozière}, \bibinfo{person}{Jonas Gehring}, \bibinfo{person}{Fabian Gloeckle}, \bibinfo{person}{Sten Sootla}, \bibinfo{person}{Itai Gat}, \bibinfo{person}{Xiaoqing~Ellen Tan}, \bibinfo{person}{Yossi Adi}, \bibinfo{person}{Jingyu Liu}, \bibinfo{person}{Romain Sauvestre}, \bibinfo{person}{Tal Remez}, \bibinfo{person}{Jérémy Rapin}, \bibinfo{person}{Artyom Kozhevnikov}, \bibinfo{person}{Ivan Evtimov}, \bibinfo{person}{Joanna Bitton}, \bibinfo{person}{Manish Bhatt}, \bibinfo{person}{Cristian~Canton Ferrer}, \bibinfo{person}{Aaron Grattafiori}, \bibinfo{person}{Wenhan Xiong}, \bibinfo{person}{Alexandre Défossez}, \bibinfo{person}{Jade Copet}, \bibinfo{person}{Faisal Azhar}, \bibinfo{person}{Hugo Touvron}, \bibinfo{person}{Louis Martin}, \bibinfo{person}{Nicolas Usunier}, \bibinfo{person}{Thomas Scialom}, {and} \bibinfo{person}{Gabriel Synnaeve}.} \bibinfo{year}{2024}\natexlab{}.
\newblock \showarticletitle{Code Llama: Open Foundation Models for Code}.
\newblock \bibinfo{journal}{\emph{arXiv preprint arXiv:2308.12950}} (\bibinfo{year}{2024}).
\newblock


\bibitem[Schieffer et~al\mbox{.}(2024)]%
        {superchip-benchmark}
\bibfield{author}{\bibinfo{person}{Gabin Schieffer}, \bibinfo{person}{Jacob Wahlgren}, \bibinfo{person}{Jie Ren}, \bibinfo{person}{Jennifer Faj}, {and} \bibinfo{person}{Ivy Peng}.} \bibinfo{year}{2024}\natexlab{}.
\newblock \showarticletitle{{Harnessing Integrated CPU-GPU System Memory for HPC: a first look into Grace Hopper}}.
\newblock \bibinfo{journal}{\emph{arXiv preprint arXiv:2407.07850}} (\bibinfo{year}{2024}).
\newblock


\bibitem[Services(2023)]%
        {aws-gh200}
\bibfield{author}{\bibinfo{person}{Amazon~Web Services}.} \bibinfo{year}{2023}\natexlab{}.
\newblock \bibinfo{title}{NVIDIA Announces Availability of NVIDIA GH200 Grace Hopper Superchips on Amazon EC2}.
\newblock \bibinfo{howpublished}{\url{https://aws.amazon.com/solutions/case-studies/nvidia-keynote-aws-reinvent-2023}}.
\newblock


\bibitem[Shoeybi et~al\mbox{.}(2020)]%
        {megatron-lm}
\bibfield{author}{\bibinfo{person}{Mohammad Shoeybi}, \bibinfo{person}{Mostofa Patwary}, \bibinfo{person}{Raul Puri}, \bibinfo{person}{Patrick LeGresley}, \bibinfo{person}{Jared Casper}, {and} \bibinfo{person}{Bryan Catanzaro}.} \bibinfo{year}{2020}\natexlab{}.
\newblock \showarticletitle{{Megatron-LM: Training Multi-Billion Parameter Language Models Using Model Parallelism}}.
\newblock \bibinfo{journal}{\emph{arXiv preprint arXiv:1909.08053}} (\bibinfo{year}{2020}).
\newblock


\bibitem[Singhal et~al\mbox{.}(2023)]%
        {medpalm}
\bibfield{author}{\bibinfo{person}{Karan Singhal}, \bibinfo{person}{Shekoofeh Azizi}, \bibinfo{person}{Tao Tu}, \bibinfo{person}{S.~Sara Mahdavi}, \bibinfo{person}{Jason Wei}, \bibinfo{person}{Hyung~Won Chung}, \bibinfo{person}{Nathan Scales}, \bibinfo{person}{Ajay Tanwani}, \bibinfo{person}{Heather Cole-Lewis}, \bibinfo{person}{Stephen Pfohl}, \bibinfo{person}{Perry Payne}, \bibinfo{person}{Martin Seneviratne}, \bibinfo{person}{Paul Gamble}, \bibinfo{person}{Chris Kelly}, \bibinfo{person}{Nathaneal Scharli}, \bibinfo{person}{Aakanksha Chowdhery}, \bibinfo{person}{Philip Mansfield}, \bibinfo{person}{Blaise Aguera~y Arcas}, \bibinfo{person}{Dale Webster}, \bibinfo{person}{Greg~S. Corrado}, \bibinfo{person}{Yossi Matias}, \bibinfo{person}{Katherine Chou}, \bibinfo{person}{Juraj Gottweis}, \bibinfo{person}{Nenad Tomasev}, \bibinfo{person}{Yun Liu}, \bibinfo{person}{Alvin Rajkomar}, \bibinfo{person}{Joelle Barral}, \bibinfo{person}{Christopher Semturs}, \bibinfo{person}{Alan Karthikesalingam}, {and}
  \bibinfo{person}{Vivek Natarajan}.} \bibinfo{year}{2023}\natexlab{}.
\newblock \showarticletitle{Large Language Models Encode Clinical Knowledge}.
\newblock \bibinfo{journal}{\emph{arXiv preprint arXiv:2212.13138}} (\bibinfo{year}{2023}).
\newblock


\bibitem[team(2023)]%
        {bloom}
\bibfield{author}{\bibinfo{person}{BigScience team}.} \bibinfo{year}{2023}\natexlab{}.
\newblock \showarticletitle{{BLOOM: A 176B-Parameter Open-Access Multilingual Language Model}}.
\newblock \bibinfo{journal}{\emph{arXiv preprint arXiv:2211.05100}} (\bibinfo{year}{2023}).
\newblock


\bibitem[Team(2024)]%
        {googlegemini}
\bibfield{author}{\bibinfo{person}{Gemini Team}.} \bibinfo{year}{2024}\natexlab{}.
\newblock \showarticletitle{{Gemini: A Family of Highly Capable Multimodal Models}}.
\newblock \bibinfo{journal}{\emph{arXiv preprint arXiv:2312.11805}} (\bibinfo{year}{2024}).
\newblock


\bibitem[{TOP500}(2024)]%
        {alps}
\bibfield{author}{\bibinfo{person}{{TOP500}}.} \bibinfo{year}{2024}\natexlab{}.
\newblock \bibinfo{title}{Alps Supercomputer}.
\newblock \bibinfo{howpublished}{\url{https://top500.org/system/180259}}.
\newblock


\bibitem[Touvron et~al\mbox{.}(2023)]%
        {llama}
\bibfield{author}{\bibinfo{person}{Hugo Touvron}, \bibinfo{person}{Thibaut Lavril}, \bibinfo{person}{Gautier Izacard}, \bibinfo{person}{Xavier Martinet}, \bibinfo{person}{Marie-Anne Lachaux}, \bibinfo{person}{Timothée Lacroix}, \bibinfo{person}{Baptiste Rozière}, \bibinfo{person}{Naman Goyal}, \bibinfo{person}{Eric Hambro}, \bibinfo{person}{Faisal Azhar}, \bibinfo{person}{Aurelien Rodriguez}, \bibinfo{person}{Armand Joulin}, \bibinfo{person}{Edouard Grave}, {and} \bibinfo{person}{Guillaume Lample}.} \bibinfo{year}{2023}\natexlab{}.
\newblock \showarticletitle{{LLaMA: Open and Efficient Foundation Language Models}}.
\newblock \bibinfo{journal}{\emph{arXiv preprint arXiv:2302.13971}} (\bibinfo{year}{2023}).
\newblock


\bibitem[Vaswani et~al\mbox{.}(2017)]%
        {transformer}
\bibfield{author}{\bibinfo{person}{Ashish Vaswani}, \bibinfo{person}{Noam Shazeer}, \bibinfo{person}{Niki Parmar}, \bibinfo{person}{Jakob Uszkoreit}, \bibinfo{person}{Llion Jones}, \bibinfo{person}{Aidan~N. Gomez}, \bibinfo{person}{Lukasz Kaiser}, {and} \bibinfo{person}{Illia Polosukhin}.} \bibinfo{year}{2017}\natexlab{}.
\newblock \showarticletitle{{Attention is All you Need}}. In \bibinfo{booktitle}{\emph{Proceedings of the 31st Conference on Neural Information Processing Systems (NIPS'17)}}.
\newblock


\bibitem[Wang et~al\mbox{.}(2018)]%
        {superneurons}
\bibfield{author}{\bibinfo{person}{Linnan Wang}, \bibinfo{person}{Jinmian Ye}, \bibinfo{person}{Yiyang Zhao}, \bibinfo{person}{Wei Wu}, \bibinfo{person}{Ang Li}, \bibinfo{person}{Shuaiwen~Leon Song}, \bibinfo{person}{Zenglin Xu}, {and} \bibinfo{person}{Tim Kraska}.} \bibinfo{year}{2018}\natexlab{}.
\newblock \showarticletitle{{SuperNeurons: Dynamic GPU Memory Management for Training Deep Neural Networks}}. In \bibinfo{booktitle}{\emph{Proceedings of the 23rd ACM SIGPLAN Symposium on Principles and Practice of Parallel Programming (PPoPP'18)}}.
\newblock


\bibitem[{xAI}(2024)]%
        {grok}
\bibfield{author}{\bibinfo{person}{{xAI}}.} \bibinfo{year}{2024}\natexlab{}.
\newblock \bibinfo{title}{Grok-2}.
\newblock \bibinfo{howpublished}{\url{https://x.ai/blog/grok-2}}.
\newblock


\bibitem[Xu et~al\mbox{.}(2024)]%
        {pie}
\bibfield{author}{\bibinfo{person}{Yi Xu}, \bibinfo{person}{Ziming Mao}, \bibinfo{person}{Xiangxi Mo}, \bibinfo{person}{Shu Liu}, {and} \bibinfo{person}{Ion Stoica}.} \bibinfo{year}{2024}\natexlab{}.
\newblock \showarticletitle{{Pie: Pooling CPU Memory for LLM Inference}}.
\newblock \bibinfo{journal}{\emph{arXiv preprint arXiv:2411.09317}} (\bibinfo{year}{2024}).
\newblock


\bibitem[Zhang et~al\mbox{.}(2024)]%
        {instruct-tuning}
\bibfield{author}{\bibinfo{person}{Shengyu Zhang}, \bibinfo{person}{Linfeng Dong}, \bibinfo{person}{Xiaoya Li}, \bibinfo{person}{Sen Zhang}, \bibinfo{person}{Xiaofei Sun}, \bibinfo{person}{Shuhe Wang}, \bibinfo{person}{Jiwei Li}, \bibinfo{person}{Runyi Hu}, \bibinfo{person}{Tianwei Zhang}, \bibinfo{person}{Fei Wu}, {and} \bibinfo{person}{Guoyin Wang}.} \bibinfo{year}{2024}\natexlab{}.
\newblock \showarticletitle{{Instruction Tuning for Large Language Models: A Survey}}.
\newblock \bibinfo{journal}{\emph{arXiv preprint arXiv:2308.10792}} (\bibinfo{year}{2024}).
\newblock


\bibitem[Zhang et~al\mbox{.}(2022)]%
        {opt}
\bibfield{author}{\bibinfo{person}{Susan Zhang}, \bibinfo{person}{Stephen Roller}, \bibinfo{person}{Naman Goyal}, \bibinfo{person}{Mikel Artetxe}, \bibinfo{person}{Moya Chen}, \bibinfo{person}{Shuohui Chen}, \bibinfo{person}{Christopher Dewan}, \bibinfo{person}{Mona Diab}, \bibinfo{person}{Xian Li}, \bibinfo{person}{Xi~Victoria Lin}, \bibinfo{person}{Todor Mihaylov}, \bibinfo{person}{Myle Ott}, \bibinfo{person}{Sam Shleifer}, \bibinfo{person}{Kurt Shuster}, \bibinfo{person}{Daniel Simig}, \bibinfo{person}{Punit~Singh Koura}, \bibinfo{person}{Anjali Sridhar}, \bibinfo{person}{Tianlu Wang}, {and} \bibinfo{person}{Luke Zettlemoyer}.} \bibinfo{year}{2022}\natexlab{}.
\newblock \showarticletitle{{OPT: Open Pre-trained Transformer Language Models}}.
\newblock \bibinfo{journal}{\emph{arXiv preprint arXiv:2205.01068}} (\bibinfo{year}{2022}).
\newblock


\bibitem[Zhang and You(2024)]%
        {speedloader}
\bibfield{author}{\bibinfo{person}{Yiqi Zhang} {and} \bibinfo{person}{Yang You}.} \bibinfo{year}{2024}\natexlab{}.
\newblock \showarticletitle{{SpeedLoader: An I/O Efficient Scheme for Heterogeneous and Distributed LLM Operation}}. In \bibinfo{booktitle}{\emph{Proceedings of the 38th Conference on Neural Information Processing Systems (NIPS'24)}}.
\newblock


\bibitem[Zhao et~al\mbox{.}(2023)]%
        {fsdp}
\bibfield{author}{\bibinfo{person}{Yanli Zhao}, \bibinfo{person}{Andrew Gu}, \bibinfo{person}{Rohan Varma}, \bibinfo{person}{Liang Luo}, \bibinfo{person}{Chien-Chin Huang}, \bibinfo{person}{Min Xu}, \bibinfo{person}{Less Wright}, \bibinfo{person}{Hamid Shojanazeri}, \bibinfo{person}{Myle Ott}, \bibinfo{person}{Sam Shleifer}, \bibinfo{person}{Alban Desmaison}, \bibinfo{person}{Can Balioglu}, \bibinfo{person}{Pritam Damania}, \bibinfo{person}{Bernard Nguyen}, \bibinfo{person}{Geeta Chauhan}, \bibinfo{person}{Yuchen Hao}, \bibinfo{person}{Ajit Mathews}, {and} \bibinfo{person}{Shen Li}.} \bibinfo{year}{2023}\natexlab{}.
\newblock \showarticletitle{{PyTorch FSDP: Experiences on Scaling Fully Sharded Data Parallel}}.
\newblock \bibinfo{journal}{\emph{arXiv preprint arXiv:2304.11277}} (\bibinfo{year}{2023}).
\newblock


\end{thebibliography}

\newpage
\clearpage
\appendix

\section{Model Config}
\label{sec:appendix:model-config}
\begin{table}[h]
    \centering
    \caption{\textbf{Model configuration in evaluation.}}
    \begin{tabular}{|c|c|c|}
        \hline
        \textbf{\# params} & \textbf{\# layer} & \textbf{Hidden size} \\
        \hline
        1, 2, 3 billion & 20, 40, 60 & 2048 \\
        4 billion & 64 & 2304 \\
        5, 6, 8 billion & 44, 53, 72 & 3072 \\
        10, 11 billion & 50, 55 & 4096 \\
        12, 13 billion & 60, 65 & 4096 \\
        15 billion & 78 & 4096 \\
        20, 25, 50, 60 billion & 25, 30, 60, 75 & 8192 \\
        70, 80 billion & 87, 100 & 8192 \\
        150, 200 billion & 45, 60  & 16384 \\
        \hline
    \end{tabular}
    \label{tab:model_config}
    % \minjia{TODO: Update the figure to be consistent with the results. In some figures, there are models with 200B parameters but are not included in this table. Also, some rows can be combined to save space, e.g., 10,11,12,13,15B have the same hidden dim but with different layers.}
\end{table}

\section{Baseline}
\label{sec:appendix:baseline}
We compare the effectiveness of \superoffload with state-of-the-art multi-billion parameter training solutions:
\begin{itemize}
    \item \textbf{PyTorch DDP}~\cite{ddp}: The standard PyTorch Transformer implementation using DistributedDataParallel (DDP) for data parallelism across multiple GPUs.
    \item \textbf{Megatron}~\cite{megatron-lm}: A widely used large-scale model training solution that employs model parallelism to efficiently scale Transformer models.
    \item \textbf{ZeRO-2}~\cite{zero}: An extension of data parallelism that eliminates memory redundancies across multiple GPUs by sharding gradients and optimizer states.
    \item \textbf{ZeRO-3}~\cite{zero}: An enhancement over ZeRO-2 that further shards model parameters across GPUs.
    \item \textbf{\zerooffload}~\cite{zero-offload}: Built on top of ZeRO-2, \zerooffload further offloads gradients and optimizer states to CPU memory, providing an efficient solution for training large-scale models with limited GPU memory. It has been adopted in DeepSpeed as the default offloading solution to CPU.
    \item \textbf{ZeRO-Infinity}~\cite{zero-infinity}: An extension of ZeRO-3 and ZeRO-Offload that enables training very large models by utilizing GPU, CPU memory, and NVMe storage through optimized memory management, prefetching, and computation-communication overlapping. For fair comparison, we only enable its CPU offloading capabilities, excluding NVMe storage offloading.
    \item \textbf{FSDP-CPU Offload}~\cite{fsdp}: Fully Sharded Data Parallel (FSDP) with CPU offloading is an advanced memory-saving technique that extends FSDP by offloading model states, including parameters, gradients, and optimizer states, to CPU when not actively needed.
\end{itemize}

\end{document}